%% file: main.tex
\def\year{2020}\relax
\documentclass[letterpaper]{article} 
\usepackage{aaai20}  
\usepackage{times}  
\usepackage{helvet} 
\usepackage{courier}  
\usepackage[hyphens]{url}  
\usepackage{graphicx} 
\usepackage{graphicx}  
\usepackage{dsfont}
\usepackage{subfig}

\title{Atari-HEAD: Atari Human Eye-Tracking and Demonstration Dataset}
\author{Ruohan Zhang,\textsuperscript{\rm 1*} Calen Walshe,\textsuperscript{\rm 2} Zhuode Liu,\textsuperscript{\rm 1} Lin Guan,\textsuperscript{\rm 1} Karl S. Muller,\textsuperscript{\rm 2} \\
\Large\textbf{Jake A. Whritner,\textsuperscript{\rm 2} Luxin Zhang,\textsuperscript{\rm 3} Mary M. Hayhoe,\textsuperscript{\rm 2} Dana H. Ballard,\textsuperscript{\rm 1,2}}\\
\textsuperscript{\rm 1} Department of Computer Science, University of Texas at Austin\\
\textsuperscript{\rm 2} Center for Perceptual Systems, University of Texas at Austin\\
\textsuperscript{\rm 3} The Robotics Institute, Carnegie Mellon University\\
\textsuperscript{*}zharu@utexas.edu
}

\begin{document}

\maketitle

\begin{abstract}
Large-scale public datasets have been shown to benefit research in multiple areas of modern artificial intelligence. For decision-making research that requires human data, high-quality datasets serve as important benchmarks to facilitate the development of new methods by providing a common reproducible standard. Many human decision-making tasks require visual attention to obtain high levels of performance. Therefore, measuring eye movements can provide a rich source of information about the strategies that humans use to solve decision-making tasks. Here, we provide a large-scale, high-quality dataset of human actions with simultaneously recorded eye movements while humans play Atari video games. The dataset consists of 117 hours of gameplay data from a diverse set of 20 games, with 8 million action demonstrations and 328 million gaze samples. We introduce a novel form of gameplay, in which the human plays in a semi-frame-by-frame manner. This leads to near-optimal game decisions and game scores that are comparable or better than known human records. We demonstrate the usefulness of the dataset through two simple applications: predicting human gaze and imitating human demonstrated actions. The quality of the data leads to promising results in both tasks. Moreover, using a learned human gaze model to inform imitation learning leads to an 115\% increase in game performance. We interpret these results as highlighting the importance of incorporating human visual attention in models of decision making and demonstrating the value of the current dataset to the research community. We hope that the scale and quality of this dataset can provide more opportunities to researchers in the areas of visual attention, imitation learning, and reinforcement learning. 
\end{abstract}

\input{1-intro.tex}
\input{2-related.tex}

\input{3-method.tex}

\input{4-gaze.tex}

\input{5-il.tex}
\input{6-discuss.tex}

\begin{small}
\bibliographystyle{aaai}
\bibliography{main.bbl}
\end{small}

\input{8-appendix-attached.tex}

\end{document}

%% file: 1-intro.tex
\section{Introduction}
\begin{figure*}
\centering
\includegraphics[width=1\textwidth]{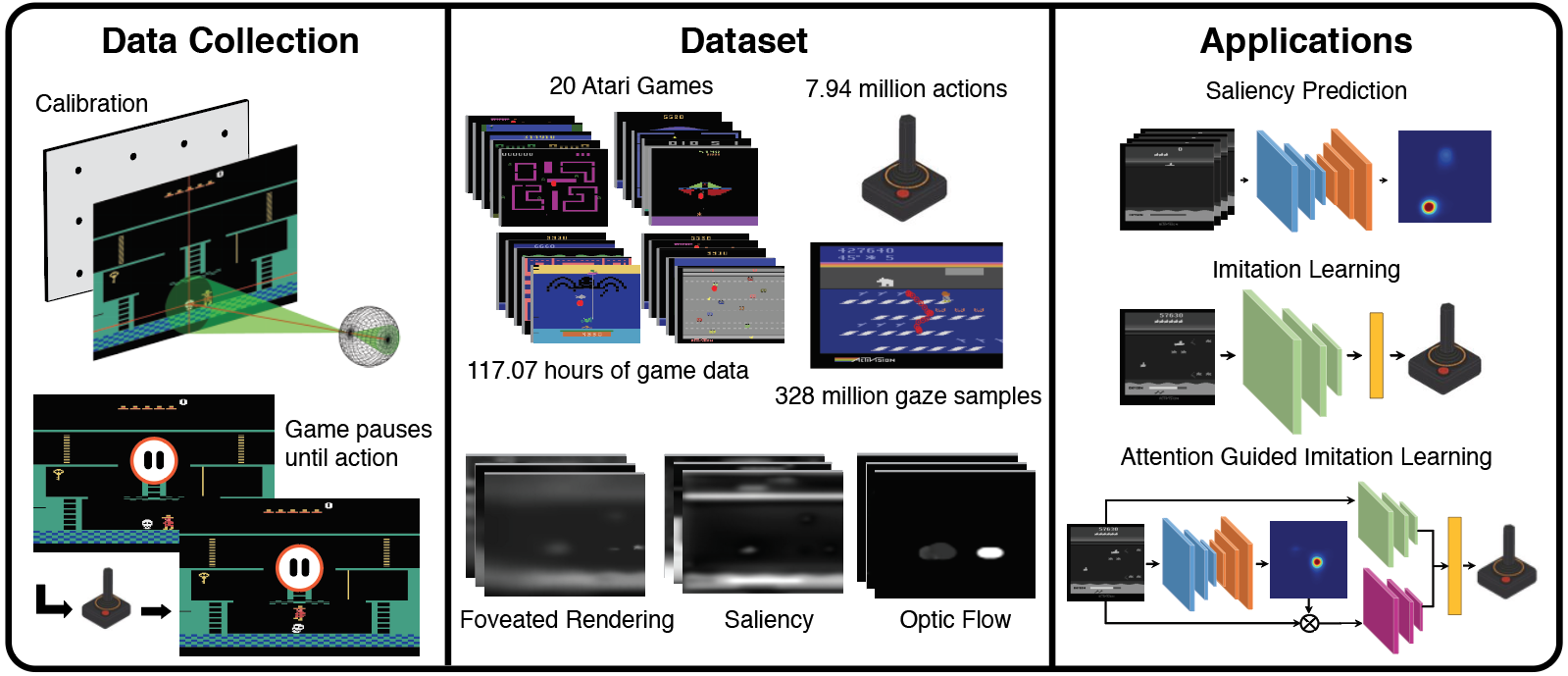}
\caption{Project schematic for the Atari-HEAD dataset.}
\label{fig:main_design}
\vspace{-0.5cm}
\end{figure*}
In modern machine learning, large-scale datasets such as ImageNet~\cite{deng2009imagenet} have played an important role in driving research progress. These datasets provide standardized benchmarks that ensure a fair comparison between algorithms.
Recently, imitation learning (IL) and reinforcement learning (RL) have achieved great success in training learning agents to solve sequential decision tasks. The goal for the learning agents is to learn a \emph{policy}--a mapping from states to actions--that maximizes long-term cumulative reward. The policy can be learned through trial and error (RL) or from an expert's demonstration (IL). A major issue of RL is its sample inefficiency and human demonstration has been shown to speed up learning~\cite{silver2016mastering,hester2018deep,de2018pre,zhangleveraging}.

However, IL results are difficult to reproduce since researchers often collect their own human demonstration data. During this process, many factors are uncontrolled -- such as individual expertise, experimental setup, data collection tools, dataset size, and experimenter bias. A publicly available dataset would greatly reduce data collection efforts and allow algorithms to be compared with the same standard. Another concern with IL is the quality of the demonstration data. For supervised IL approaches like behavior cloning, learning from sub-optimal demonstration can result in poor performance. Therefore the quality of human demonstration must be ensured.

Visual perception is a key challenge in modern RL and IL research due to a high-dimensional state space (e.g., raw images) as input. Humans face the same problem when performing all kinds of visuomotor tasks in daily life. One intelligent mechanism that has evolved in humans, but has not yet been fully developed for machines, is visual attention -- the ability to identify, process, and respond to a reduced set of important features of visual input. This powerful feature extraction mechanism, if learned by AIs, could make learning more efficient.

Human overt attention is revealed by eye movements (gaze). In complex tasks, human eye movements are used by the visual system to a) identify structures in the environment that are critical for solving the task and b) exploit those structures by moving the high resolution part of the visual field (fovea) to those locations via eye movements. Considerable evidence has shown that human gaze can be considered as an overt behavioral signal that encodes a wealth of information about both the motivation behind an action and the anticipated reward of an action~\cite{hayhoe2005eye,hayhoe2014modeling,johnson2014predicting,zhang2018modeling}. Recent work has also proposed learning visual attention from human gaze as an intermediate step towards learning the decision policy, and this intermediate signal has been shown to improve policy learning~\cite{li2018eye,zhang2018agil,xia2019periphery,chen2019gaze,liu2019gaze,deng2019drivers}

Addressing the demands and challenges described above, we collected a large-scale dataset of humans playing Atari video games -- one of the most widely used task domain in RL and IL research. The dataset is named Atari-HEAD (\textbf{Atari} \textbf{H}uman \textbf{E}ye-Tracking \textbf{A}nd \textbf{D}emonstration)\footnote{Available at: https://zenodo.org/record/2603190}. An overview of this project can be found in Fig.~\ref{fig:main_design}. In collecting Atari-HEAD, we strictly follow standard data collection protocols for human studies and designed a special method to ensure the quality of demonstration policies. The result of these efforts is a dataset with expert-level task performance and minimal recording error. Making this dataset publicly available saves the effort of data collection and provides a benchmark for researchers who use Atari games as their task domain. Having both action and gaze data enables research that aims at bridging attention and control.

%% file: 2-related.tex
\section{Related Work}
In imitation learning research, the Atari Grand Challenge dataset pioneered the effort of collecting a large-scale public dataset of Atari games~\cite{kurin2017atari}. The human demonstration was collected through online crowdsourcing with players of diverse skill levels. Recently, researchers have spent significant effort in building large-scale datasets of human demonstrations in various tasks, including driving~\cite{yu2018bdd100k}, playing Minecraft~\cite{guss2019minerl}, and manipulating simulated robots~\cite{mandlekar2018roboturk}. Our dataset joins their effort in providing a standard dataset for the RL and IL research community. 

Gaze prediction was formalized as a visual saliency prediction problem in computer vision research~\cite{itti1998model}. Large-scale datasets have enabled deep learning approaches to make tremendous advances in this area. Examples include MIT saliency benchmark~\cite{mit-saliency-benchmark}, CAT2000~\cite{borji2015cat2000}, and SALICON~\cite{jiang2015salicon}. However, the traditional saliency prediction task does not involve tasks nor human decisions. The humans look at static images or videos in a free-viewing manner without performing any particular task, and only the eye movements are recorded and modeled. How humans distribute their visual attention for dynamic, reward-seeking visuomotor tasks has received less attention in research on saliency. Recently, eye-tracking video datasets of subjects cooking~\cite{li2018eye} and driving~\cite{alletto2016dr} in naturalistic environment have been published; this allows researchers to study the relation between attention and decision. We hope that the Atari-HEAD dataset can serve a similar purpose for visual saliency and visuomotor behavior research.


%% file: 3-method.tex
\section{Atari-Head: Design and Data Collection}
Our data was collected using the Arcade Learning Environment (ALE)~\cite{bellemare2012arcade}. These games capture many interesting aspects of the natural visuomotor tasks while allowing better experimental control than real-world tasks. ALE is deterministic given the same game seed. While collecting human data, the seed is randomly generated to introduce stochasticity for gameplay. We pick 20 games that span a variety of dynamics, visual features, reward mechanisms, and difficulty levels (for both human and AI). Game images along with eye tracking data can be found in Appendix Fig 1.  

For every game image frame $i$, we recorded its corresponding image frame $I_i$, human keystroke action $a_i$, human decision time $t_i$, gaze positions $g_{i1}...g_{in}$, and the immediate reward $r_i$ returned by the environment. The gaze data was recorded using an EyeLink 1000 eye tracker at 1000Hz. The game screen was $64.6 \times 40.0$cm (or 1280$\times$840 in pixels), and the distance to the subjects' eyes was $78.7$cm. The visual angle of an object is a measure of the size of the object's image on the retina. The visual angle of the screen was $44.6 \times 28.5$ visual degrees. 

The human subjects were amateur players who were familiar with the games. The human research was approved by the University of Texas at Austin Institutional Review Board with approval number 2006-06-0085. We collected data from 4 subjects playing 20 games. The total collected game time is 117.07 hours, with 7,937,159 action demonstrations and 328,870,044 usable gaze samples.

The subjects were only allowed to play for 15 minutes, and were required to rest for at least 15 minutes before the next trial. We mainly collected human data from the first 15 minutes of game play, since for most games AIs have not reached human performance at a 15-minute cutoff. Therefore we reset the game to start from the beginning for every trial. However, it is also interesting to know the human performance limit, hence for each game we let one human player play until the game terminated, or a 2 hour maximum time limit has been reached.

\paragraph{Eye-tracking accuracy}
The Eyelink 1000 tracker was calibrated using a 16-point calibration procedure at the beginning of each trial, and the same 16 points were used at the end of trial to estimate the gaze positional error. The average end-of-trial gaze positional error across 471 trials was 0.40 visual degrees (or 2.94pixels/0.56cm), less than 1\% of the stimulus size. Such high tracking accuracy is critical for Atari games, since many task-relevant objects are small.


\paragraph{Semi-frame-by-frame game mode}

\begin{figure}
\centering
\subfloat[No-op]{\includegraphics[width=0.22\textwidth]{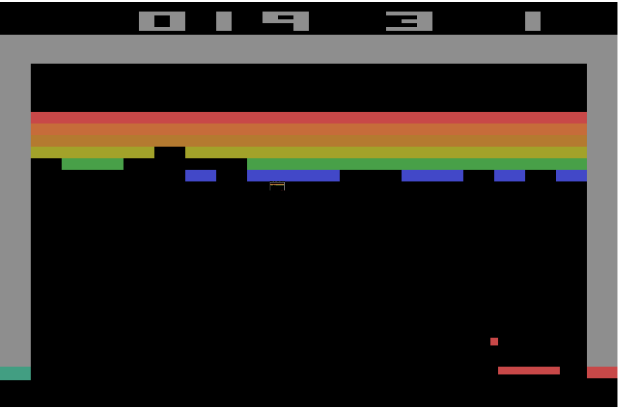}}
\subfloat[Left]{\includegraphics[width=0.22025\textwidth]{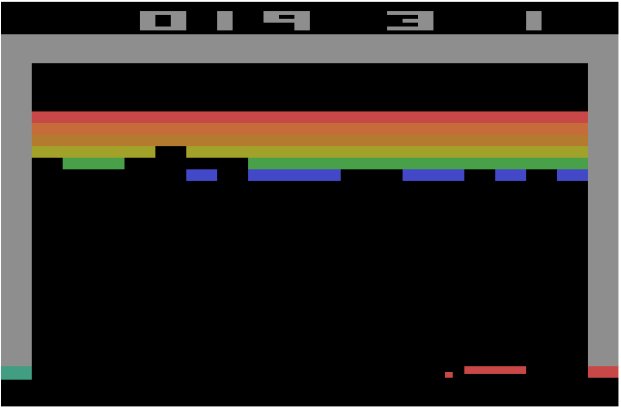}}
\caption{State-action mismatch (game Breakout, game speed at 60Hz). The state $s_0$ at time $t_0$ is shown in (a), the correct action would be to move the paddle left to catch the ball. However, due to the human player's delayed reaction, that action is executed 287ms (17 frames) later, as shown in (b). This delay leads to two undesirable consequences: 1) The player loses a life in the game; 2) Action ``Left" is paired with state $s_{17}$, instead of $s_0$, which posits a serious issue for algorithms that attempts to learn the state-action mapping.}
\label{fig:mismatch}
\end{figure}

In the default ALE setting, the game runs continuously at 60Hz, a speed that is very challenging even for expert human players. Previous studies have collected human data, or evaluated human performance at this speed~\cite{mnih2015human,wang2016dueling,kurin2017atari,hester2018deep}. However, we argue that in order to build a dataset useful for algorithms such as IL, a slower speed should be used. An innovative feature of our setup is that the game pauses at every frame, until a keyboard action is taken by the human player. If desired, the subjects can hold down a key and the game will run continuously at 20Hz, a speed that is reported to be comfortable for most players. The reasons for such a setup are as follows:

\emph{Resolving state-action mismatch} Closed-loop human visuomotor reaction time $\Delta t$ is around 250-300 milliseconds. Therefore, during continuous gameplay, $s_t$ and $a_t$ that are simultaneously recorded at time step $t$ could be mismatched. Action $a_t$ could be intended for a state $s_{t-\Delta t}$ 250-300ms ago. An example that illustrates this point is shown in Fig.~\ref{fig:mismatch}. This effect causes a serious issue for supervised learning algorithms, since label $a_t$ and input $s_t$ are no longer matched. Frame-by-frame game play ensures that $s_t$ and $a_t$ are matched at every timestep. 

\begin{figure}
\centering
\subfloat[]{\includegraphics[width=0.22\textwidth]{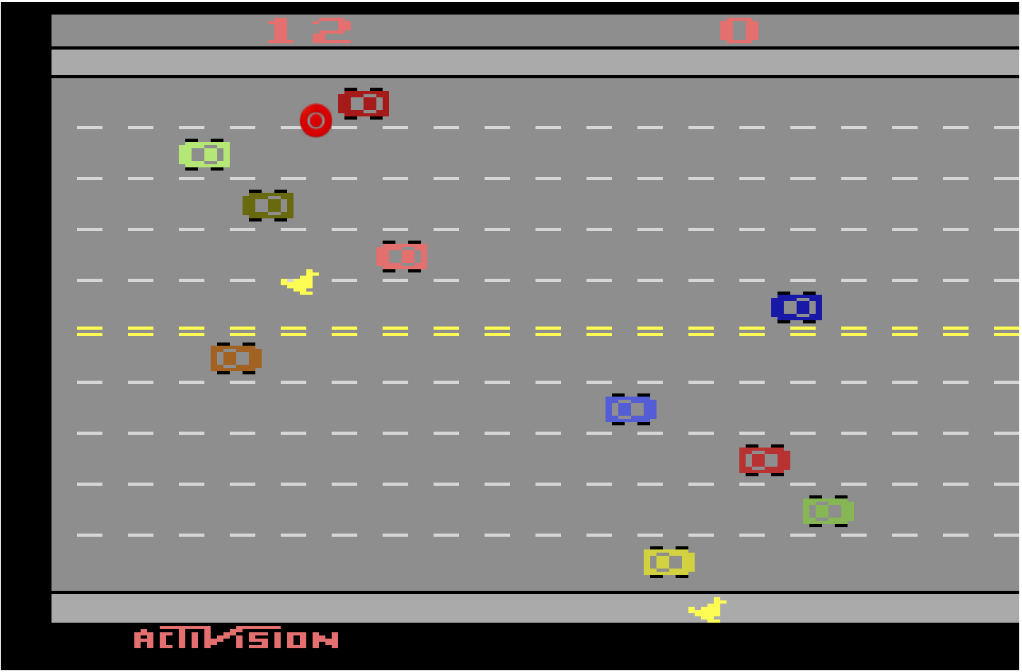}}
\subfloat[]{\includegraphics[width=0.22\textwidth]{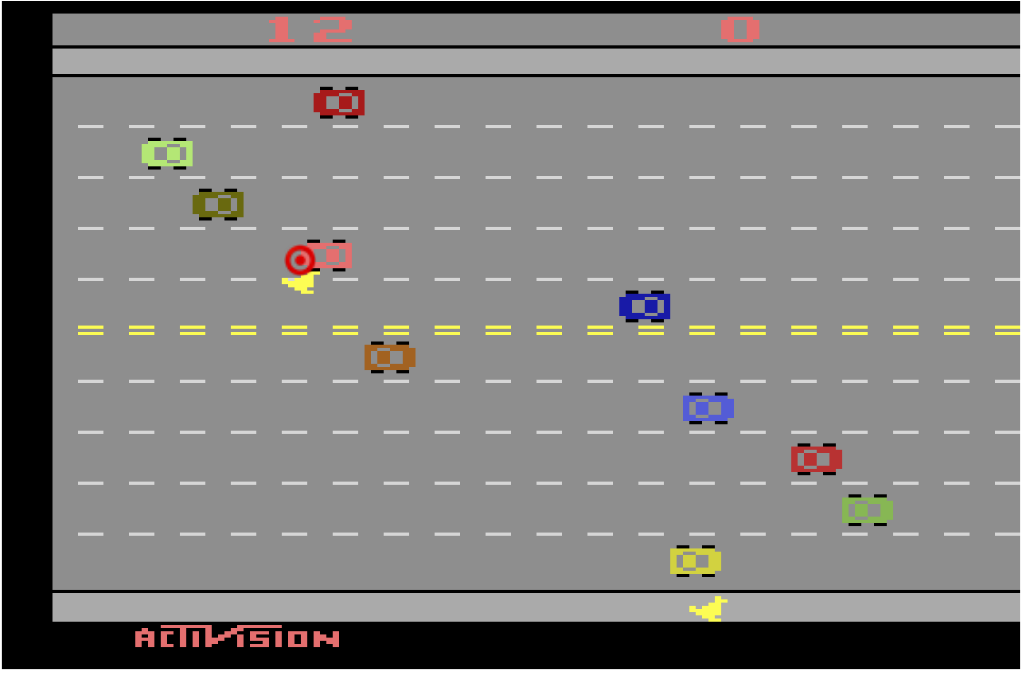}}
\caption{Inattentive blindness (game Freeway, game speed at 60Hz). (a) The player's attention (red dot) was on the red car. (b) The pink car hits the chicken controlled by the player 205ms later. Due to the fast pace of the game, the human player was not able to make an eye movement to attend and respond to the pink car.}
\label{fig:inattentive}
\end{figure}
\begin{figure}
\centering
\subfloat[217ms]{\includegraphics[width=0.22\textwidth]{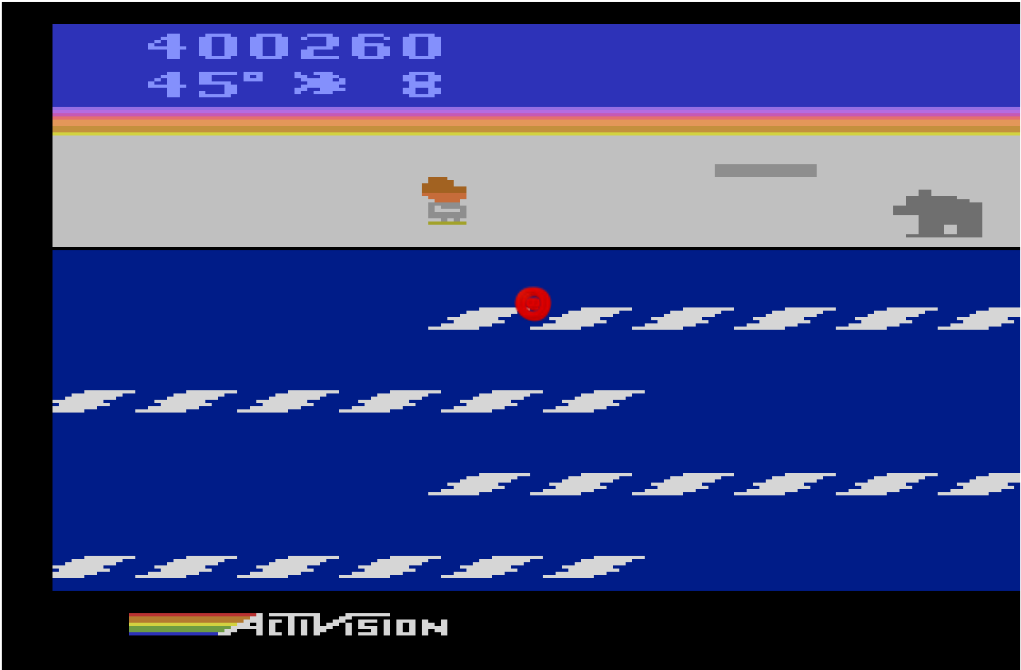}}
\subfloat[1158ms]{\includegraphics[width=0.22\textwidth]{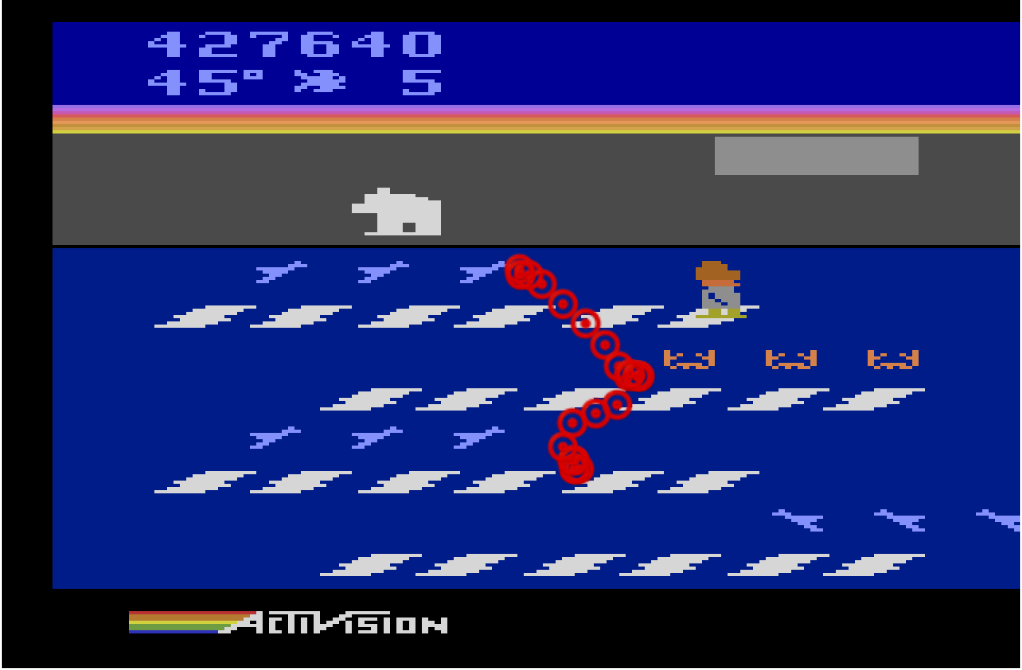}}
\caption{Scanpath and reaction time (game Frostbite, frame-by-frame mode). (a) A simple game state that only takes one fixation and 217ms to make a decision. (b) A complicated state which requires a sequence of eye movements and 1158ms to plan the next action. Our game mode allows enough time for the human player to process visual information and find the optimal action.}
\label{fig:decision_time}
\end{figure}

\emph{Maximizing human performance} Frame-by-frame mode makes gameplay more relaxing and reduces fatigue, which could normally result in blinking and would corrupt eye-tracking data. More importantly, this design reduces sub-optimal decisions caused by inattentive blindness. See Fig.~\ref{fig:inattentive} for an example. 

\emph{Highlighting critical states that require multiple eye movements} 
Human decision time and all eye movements were recorded at every frame. Hypothetically, the states that could lead to a large reward or penalty, or the ones that require sophisticated planning, will take longer and require multiple eye movements for the player to make a decision. Fig.~\ref{fig:decision_time} shows an example. Stopping gameplay means that the observer can use eye-movements to resolve complex situations like (b). This is important because if the algorithm is going to learn from eye-movements it must contain all ``relevant" eye-movements.

\begin{table*}
\centering
\scalebox{0.85}{
\begin{tabular}{c | c c c c c c c c | c | c}
\hline
& Mnih & Wang & Hester & Kurin & de la Cruz & AtariHEAD & AtariHEAD & AtariHEAD & Community & RL\\
&  &  &  & & &  15-min avg.  & 15-min best  &  2-hour & Record  \\
\hline
alien           & 6,875     & 7,127.7   & 29,160    & -       & -       & 27,923    & 34,980    & \textbf{107,140}$^\dagger$    &103,583   &9,491.7\\
asterix         & 8,503     & 8,503.3   & 18,100    & -       & 14,300  & 110,133.3 & 135,000   & \textbf{1,000,000}$\ddagger$  &\textbf{1,000,000} &428,200.3\\
bank\_heist      & 734.4     & 753.1     & 7,465     & -       & -       & 5,631.3   & 6,503     & \textbf{66,531}$^\dagger$    &47,047    &1,611.9\\
berzerk         & -         & 2,630.4   & -         & -       & -       & 6,799     & 7,950     &55,220$^\star$       &\textbf{171,770}   &2,545.6 \\
breakout        & 31.8      & 30.5      & 79        & -       & 59      & 439.7     & 554       & \textbf{864}$\ddagger$        &\textbf{864} &612.5\\
centipede       & 11,963    & 12,017    & -         & -       & -       & 45,064    & 55,932    & 415,160$^\star$      &\textbf{668,438}    &9,015.5\\
demon\_attack    & 3,401     & 3,442.8   & 6,190     & -       & -       & 7,097.3   & 10,460    & 107,045$^\star$     &108,075    &\textbf{111,185.2}\\
enduro          & 309.6     & 860.5     & 803       & -       & -       & 336.4     & 392       & \textbf{4,886}$^\star$        &-  &2,259.3\\
freeway         & 29.6      & 29.6      & 32        & -       & -       & 31.1      & 33        & 33$^\dagger$         &\textbf{34}         &\textbf{34.0}\\
frostbite       & 4,335     & 4,334.7   & -         & -       & -       & 31,731.5  & 50,630    & \textbf{453,880}$^\star$      &418,340    &9,590.5\\
hero            & 25,763    & 30,826.4  & 99,320    & -       & -       & 59,999.8  & 77,185    & 541,640$^\star$      &\textbf{1,000,000}  &55,887.4\\
montezuma       & 4,367     & 4,753.3   & 34,900    & 27,900  & -       & 38,715    & 46,000    & 270,400$^\star$      &\textbf{400,000}    &384.0\\
ms\_pacman        & 15,693    & 15,375.0  & 55,021    & 29,311  & 18,241  & 28,031    & 36,061    & 93,721$^\dagger$   &\textbf{123,200}    &6,283.5\\
name\_this\_game  & 4,076     & 8,049.0   & 19,380    & -       & 4,840   & 7,661.5   & 8,870     & \textbf{21,850}$^\dagger$   &21,210     &13,439.4\\
phoenix         & -         & 7,242.6   & -         & -       & -       & 30,800.5  & 40,780    & \textbf{485,660}$^\star$      &373,690    &108,528.6\\ 
riverraid       & 13,513    & 17,118    & 39,710    & -       & -       & 20,048    & 22,590    & 59,420$^\dagger$     &\textbf{86,520}     &-\\
road\_runner     & 7,845     & 7,845     & 20,200    & -       & -       & 78,655    & 99,400    & 99,400$^\dagger$    &\textbf{210,200}    &69,524.0\\
seaquest        & 20,182    & 42,054.7  & 101,120   & -       & -       & 52,774    & 64,710    & \textbf{585,570}$^\star$      &294,940    &50,254.2\\
space\_invaders  & 1,652     & 1,668.7   & -         & 3,355   & 1,840   & 3,527     & 5,130     & 49,340$^\star$      &\textbf{110,000}    &18,789.0\\
venture         & 1,188     & 1,187.5   & -         & -       & -       & 8,335     & 11,800    & \textbf{28,600}$^\dagger$     &-           &1,107.0\\
\hline
\hline
\end{tabular}}
\caption{A comparison of human scores for 20 Atari games across datasets. The scores reported for~\cite{hester2018deep,kurin2017atari,de2018pre} are the best human scores of each game. \citeauthor{mnih2015human} and \citeauthor{wang2016dueling} are average scores. The community world record is from Twin Galaxies, an official supplier of verified world records by Guinness World Records. Note that the display and game difficulty may vary slightly across platforms, here we try to find the game version that matches our setting to the best of our knowledge. For Atari-HEAD 2-hour performance, $^\dagger$: game terminated. $^\star$: Two-hour experiment time limit has been reached before the game terminated. If the human players continue to play, they could potentially achieve higher scores. $\ddagger$: Maximum score allowed by the game reached. Disclaimer: Our human data is recorded in the semi-frame-by-frame mode discussed above and is intended to be used for research purposes, hence should not be submitted to the gaming community for competition. }
\label{tbl:human}
\end{table*}

\paragraph{Dataset statistics}
The experimental designs result in a high-quality human demonstration dataset. The optimality of demonstrated actions can be intuitively measured by final game scores (when the players lose their last life). In Table~\ref{tbl:human}, we compare our human scores with ones reported in previous literature, along with Atari game world records, as well as one of the best RL agent's performance~\cite{hessel2018rainbow}. We reported the average and the best game scores in 15-minute trials, as well as the highest score reached in the 2-hour game play mode. The immediate observation is that our design leads to better human performance compared to those previously reported. The community world record is from Twin Galaxies\footnote{https://www.twingalaxies.com/games}, an official supplier of verified world records by Guinness World Records. For 8 games, our human players have obtained comparable or better scores than world records. For 6 other games, the 2-hour time limit was reached but the human players could surpass the world record if they continued to play. 

In recent years, the gap between human and machine performance in many tasks has substantially narrowed~\cite{mnih2015human}. AI agents such as DQN play the game in the frame-by-frame manner (although reaction time is not a big issue for RL agents), but in previous literature humans played the game continuously at 60Hz. In our case, allowing human players to have enough decision time sets a stronger human performance baseline for RL agents. Our human score statistics indicate that humans retain advantages in these games, especially ones that require multitasking and attention. For difficult games recognized by the RL research community, such as Montezuma's Revenge, human performance is still much higher than that of AI.

\begin{figure*}
\centering
\includegraphics[width=1\textwidth]{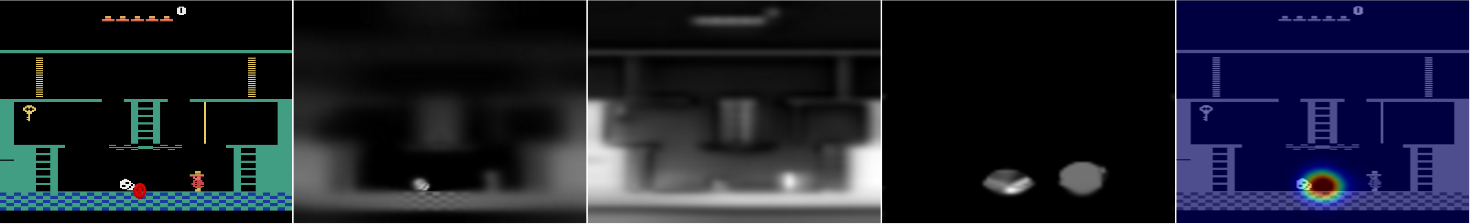}
\caption{Gaze prediction results for Montezuma's revenge. First column: game screenshots with red dots indicating the human gaze positions. Second column: biologically plausible retinal image, generated by foveated rendering algorithm~\cite{perry2002gaze}. Third column: image saliency calculated by the classic Itti-Koch saliency model~\cite{itti1998model}. Fourth column: Farnebeck optical flow, calculated using the frame in the first column and its previous frame~\cite{farneback2003two}. Fifth column: predicted gaze distribution by convolution-deconvolution network, overlayed on top of the original image. }
\label{fig:gaze}
\end{figure*}

%% file: 4-gaze.tex
\section{Applications of Atari-HEAD}
Next, we will demonstrate two main modeling tasks that can be accomplished with this dataset: learning attention and learning action from humans. We will define both tasks, discuss inputs and outputs of the models, propose evaluation metrics, show baseline modeling results, and mention potential future directions for both tasks. 
\subsection{Saliency prediction}
\paragraph{Task definition} The first learning task could be training an agent to imitate human's gaze behaviors, i.e., learning to attend to important regions of a given image. The problem is formalized as a visual saliency prediction problem in computer vision research. The problem can be formulated as:
\begin{quote}
    Given a state $s_t$, learn to predict human gaze positions $g_t$, i.e., learn $P(g|s)$.
\end{quote}

\paragraph{Inputs and outputs}
In the above formulation, note that $g_t$ could be a set of positions in our dataset. $s_t$ could be a single image $I_t$, or it could include a stack of images $I_{t-n}\dots I_t$ to take into account more history. This includes information such as motion that can make states Markovian~\cite{mnih2015human}. The images are published in RGB format, but it is common to convert them to be grayscale~\cite{mnih2015human}. Note that for this dataset, two adjacent images, actions, or gaze locations are highly correlated. We suggest that users split data first then shuffle, instead of shuffle first then split, so one can avoid putting one frame in the training set and its neighboring frame in the testing set.

In saliency prediction, additional image statistics are shown to be correlated with visual attention and useful for gaze prediction~\cite{palazzi2018predicting,li2018eye,zhang2018agil}. We also provide tools to extract optical flow~\cite{farneback2003two} and hand-crafted bottom-up saliency features (orientation and intensity)~\cite{itti1998model}. Examples of these can be seen in the third and fourth columns of Fig.~\ref{fig:gaze}. They can be directly used as reasonable guesses for gaze locations.

The gaze prediction model should output $P(g_t|s_t)$. In standard practice, discrete gaze positions are converted into a continuous distribution~\cite{bylinskii2018different} by blurring each fixation location using a Gaussian with $\sigma$ equals to one visual degree~\cite{le2013methods}. Hence the gaze prediction model will learn to predict this continuous probability distribution over the given image, which will be referred as a saliency map. 

\paragraph{Evaluation metrics} Once the conversion is done, at least eight well-known metrics can be applied to measure prediction accuracy~\cite{bylinskii2018different}. Let $P$ denote the predicted saliency map, $Q$ denote the ground truth, and $i$ denote the $i$th pixel. We discuss four selected metrics here:
\begin{itemize}
    \item \textbf{Area Under ROC Curve (AUC):} between 0 and 1. One can treat predicted saliency map as a binary classifier to indicate whether a pixel is fixated or not. Hence AUC, one of the most widely used metric in signal detection and classification problems can be applied here.
    \item \textbf{Normalized Scanpath Saliency (NSS):} This metric measures the normalized saliency at gaze positions by subtracting the mean predicted saliency value. It is sensitive to false positives and differences in saliency across predicted saliency map, but is invariant to linear transformations like contrast offsets:  
    \begin{equation}
    NSS(P,Q) = \frac{1}{\sum_i Q_i} \sum_i \Big(\frac{P_i - \mu(P)}{\sigma(P)} \times Q_i\Big)
    \end{equation}
    \item \textbf{Kullback-Leibler Divergence (KL):} This metric is widely used to measure the difference between two probability distributions. It is also differentiable hence can be used as the loss function to train neural networks: \begin{equation}
    KL(P,Q) = \sum_i Q_i \log \Big(\epsilon + \frac{Q_i}{\epsilon + P_i}\Big)
    \end{equation}
    $\epsilon$ is a small regularization constant and determines how much zero-valued predictions are penalized. KL is asymmetric and very sensitive to zero-valued predictions.
    \item \textbf{Pearson’s Correlation Coefficient (CC):} between 0 and 1. It measures the linear relationship between two distributions. 
    \begin{equation}
    CC(P,Q) = \frac{\sigma(P,Q)}{\sigma(P) \times \sigma(Q)}
    \end{equation}
    where $\sigma(P,Q)$ denotes the covariance. CC is symmetric and penalizes false positives and negatives equally. 
\end{itemize}
Note that KL and CC are distribution-based metrics, therefore the aforementioned process of converting discrete gaze positions to distributions is mandatory. However, for location-based metrics (AUC and NSS) the conversion is optional. Other usable metrics include Information Gain, Histogram Intersection, Shuffled AUC, Earth Mover’s Distance. For a comprehensive survey about their properties, please see~\citeauthor{bylinskii2018different}~\shortcite{bylinskii2018different}.

\paragraph{Baseline model and results}
We trained a convolution-deconvolution gaze network~\cite{palazzi2018predicting,zhang2018agil,zhang2018learning,deng2019drivers} with KL divergence ($\epsilon=1e-10$) as loss function to predict human gaze positions. The details of the network design can be found in Appendix Fig.~2. A separate network is trained for each game. We use 80\% data for training and 20\% for testing. 

\begin{figure}
\centering
\includegraphics[width=.45\textwidth]{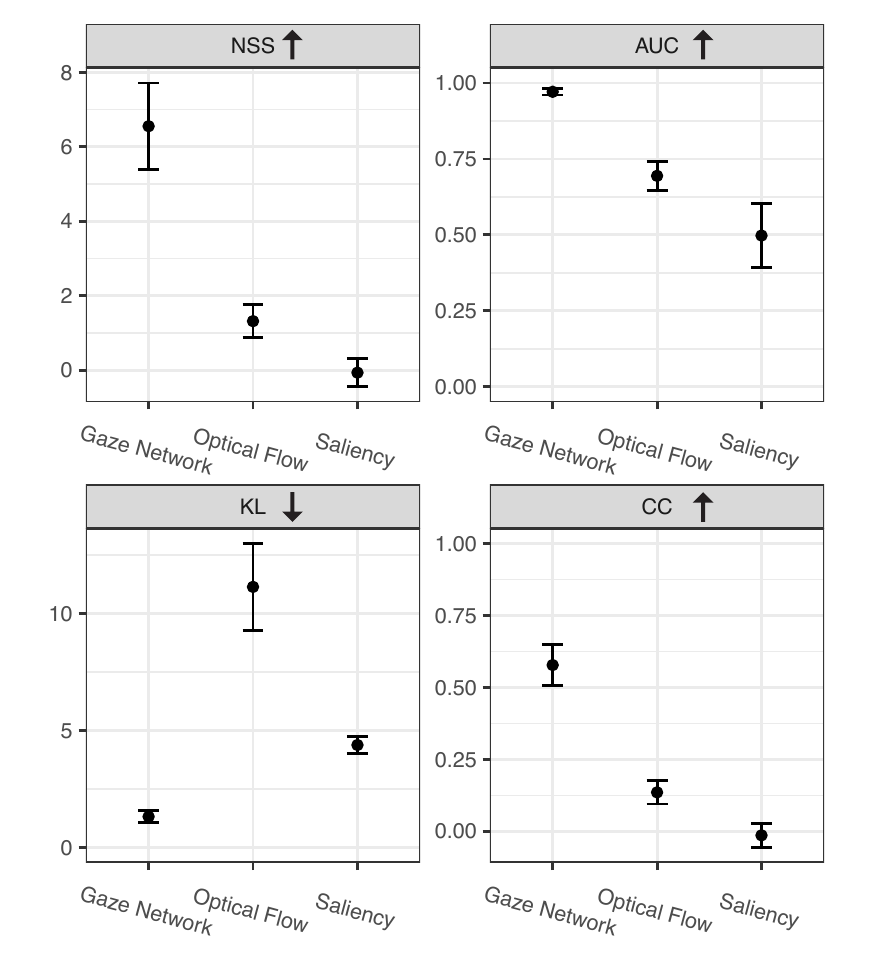}
\caption{Gaze prediction results measured using four standard metrics, averaged across 20 games. As expected, a convolution-deconvolution network (gaze network) is able to predict human gaze much more accurately than motion-based and image saliency-based models. Error bars indicate standard deviation across games (N=20). }
\label{fig:score_metrics}
\end{figure}

Aggregated modeling results can be seen in Fig.~\ref{fig:score_metrics}. As expected, the learning-based neural network model outperforms optical flow and bottom-up saliency models by a large margin in all metrics. The prediction accuracy overall is high (average AUC of 0.971), although varies across games (min AUC: 0.945-Ms.Pacman, max: 0.988-Enduro). Results for each individual game can be found in Appendix Table 1. The predicted saliency maps can be visualized in Fig.~\ref{fig:gaze} and Appendix Fig.~3. 

The saliency prediction results using the dataset are considered highly accurate in saliency prediction research. One reason is the large amount of training data available provided by the dataset. Another reason is that the chosen tasks are reward-seeking and demanding, therefore human gaze is mostly directed towards image features that are strongly associated with reward and hence become highly predictable~\cite{hayhoe2005eye,hayhoe2014modeling}.

\paragraph{Potential future work}
To further improve the prediction accuracy, researchers can optionally use additional inputs (motion, bottom-up saliency, or image semantics) along with the original images to predict human gaze positions. Several previous works have shown these signals are helpful for gaze prediction in visuomotor tasks~\cite{zhang2018agil,palazzi2018predicting,li2018eye}. Another option would be using a recurrent neural network to represent information from past frames as memory, instead of stacking multiple images~\cite{mnih2014recurrent,xia2019periphery,zelinsky2019benchmarking}. Recurrent network models also allow one to model eye movements scanpaths as a sequence prediction problem.

%% file: 5-il.tex
\subsection{Imitation learning} 
\paragraph{Task definition}
The next task is to learn from human demonstrated actions. This standard IL problem is formulated as follows:
\begin{quote}
    Given a state $s_t$, learn to predict human action $a_t$, i.e,. learn $P(a|s)$, or equivalently, policy $\pi(s,a)$.
\end{quote}
With human gaze data, we propose to use attention information to improve policy learning. This attention-guided learning problem is formulated as follows:
\begin{quote}
    Given a state $s_t$ and human gaze positions $g_t$, learn to predict human action $a_t$, i.e., learn $P(a|s,g)$.
\end{quote}
Another potential formulation is a joint learning problem:
\begin{quote}
    Given a state $s_t$, learn to jointly predict human action $a_t$ and gaze positions $g_t$, i.e., learn $P(a, g|s)$.
\end{quote}

\paragraph{Evaluation metrics}
\begin{itemize}
    \item \textbf{Behavior matching accuracy:} It measures the accuracy in predicting human actions. Since Atari games have a discrete action space (18 actions), one can treat the prediction task as a 18-way classification problem with standard log likelihood loss: 
    \begin{equation}
        J = - \sum_t^T \sum_{a=0}^{17} \mathds{1}_{a_t=a} \log P(a_t = a|s_t)
    \end{equation}
    This supervised learning approach for imitation learning is commonly referred to as behavior cloning. 
    \item \textbf{Game score:} The model that predicts human actions is effectively a gaming AI. Its performance can be directly measured by the final game score.
\end{itemize}
Note that the results on these two metrics may not necessarily be correlated, as we will show later.

\paragraph{Baseline model and results}
For standard IL, we trained a convolutional network using the classification loss above. To incorporate gaze information into IL, we use a two-channel policy network~\cite{zhang2018agil}. The policy network uses the saliency map predicted by the gaze network to mask the input image. This mask can be applied to the image to generate a ``foveated" representation of the image that highlights the attended visual features~\cite{li2018eye,zhang2017attention,zhang2018agil,xia2019periphery,chen2019gaze}. The design of both networks can be found in Appendix Fig.~4 and 5. 

\begin{figure}
\includegraphics[width=.375\textwidth]{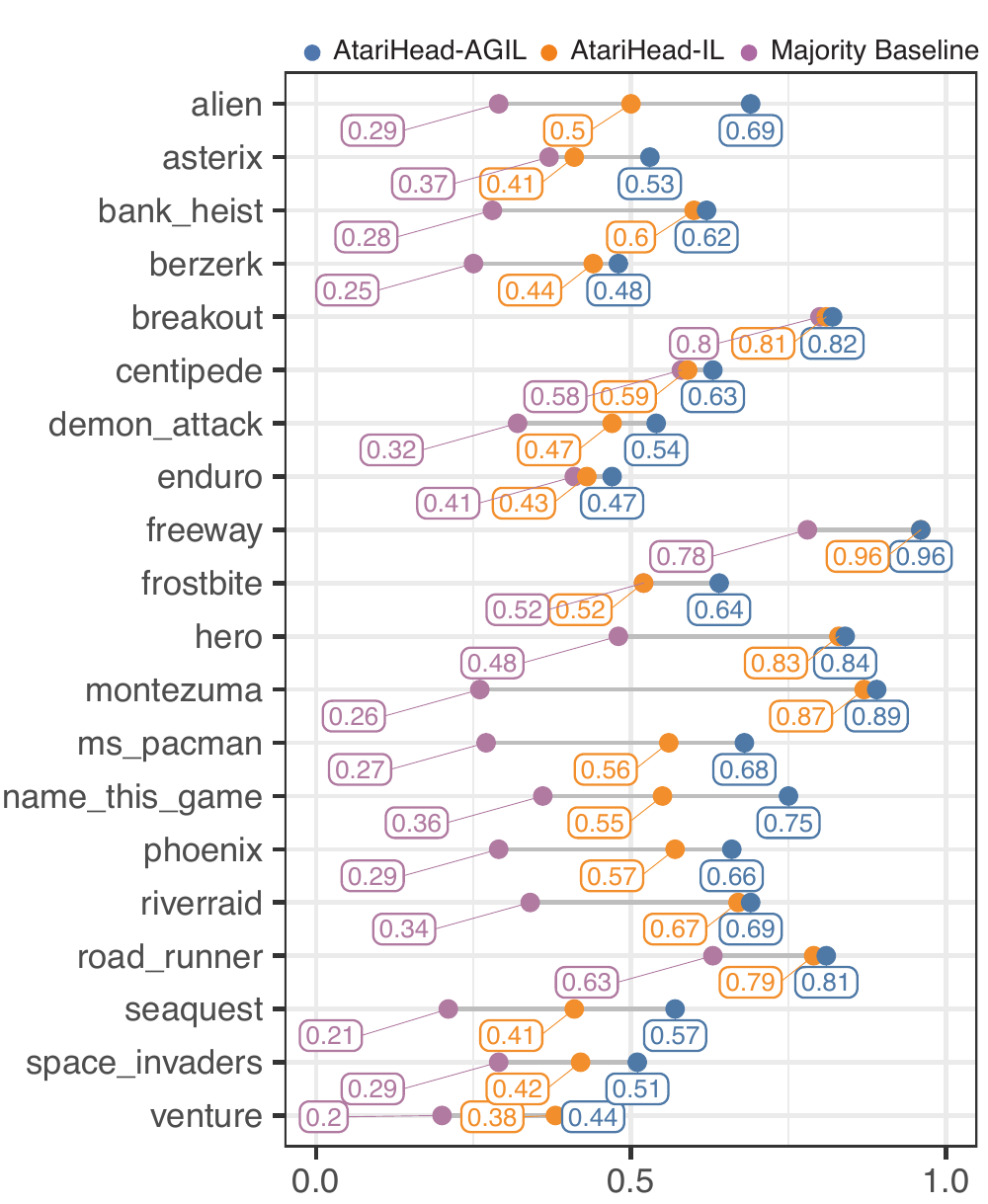}
\caption{Behavior matching accuracy of different models. Majority baseline simply predicts the majority class in that game (the most frequent action). Imitation learning: Standard behavior cloning. AGIL: policy network that includes saliency map predicted by the gaze network. Random guess accuracy: 0.06. }
\label{fig:accuracy}
\end{figure}

The performance measured by behavior matching accuracy can be seen in Fig.~\ref{fig:accuracy} and Appendix Table 2. The main result is that incorporating attention improves accuracy on all games with an average improvement of 7\%. However, the magnitude of improvement varies across games. The games with most improvements are Name This Game ($19\%$), Alien ($19\%$), Seaquest ($16\%$), Ms.Pacman ($12\%$), Asterix ($12\%$), and Frostbite ($12\%$). These are games where many task-relevant objects appear on the screen simultaneously. As a result, the current behavioral target is often ambiguous without attention information, therefore incorporating attention leads to better prediction. 

Next we look at game scores obtained by different models using different datasets, shown in Fig.~\ref{fig:gamescore} (additional statistics can be found in Appendix Table 3). We include IL (behavior cloning) results from two previous datasets~\cite{hester2018deep,kurin2017atari}. The key observation is that the scale and quality of our data results in better performance compared to these datasets. More importantly, this second metric confirms again that attention information is useful for IL. The AGIL model improves game performance on 19 games, with an average improvement of 115.26\%.

Intuitively, knowing where humans look provides useful information on what action they take. Standard IL can only capture \emph{what} the human teacher did, without knowing \emph{why} the decision was made. Visual attention is a good indicator of why a particular decision was made. Incorporating such information leads to better performance in both metrics.

\paragraph{Potential future work} 
As mentioned before, the results on the accuracy and score metrics may not necessarily be strongly correlated. For instance a 1\% increase in accuracy leads to a 1138\% improvement in scores for the game Breakout, while for Space Invaders, a 9\% increase leads to minor improvement (0.45\%) in game scores. In addition, it was found in experiments that adding dropout to the network improves human action prediction accuracy but hurts game performance, an issue worth further investigation. 

There is still a large performance gap between the learning agent and the human scores reported in Table~\ref{tbl:human}. Part of this is due to the inherent problems with behavior cloning such as covariate shift in state distribution~\cite{ross2011reduction}. Using IL and saliency prediction as auxiliary tasks~\cite{hester2018deep,jaderberg2016reinforcement,zhang2019attention} or a pre-training step for RL~\cite{de2018pre} are promising ways to improve game performance.

\begin{figure}
\includegraphics[width=.375\textwidth]{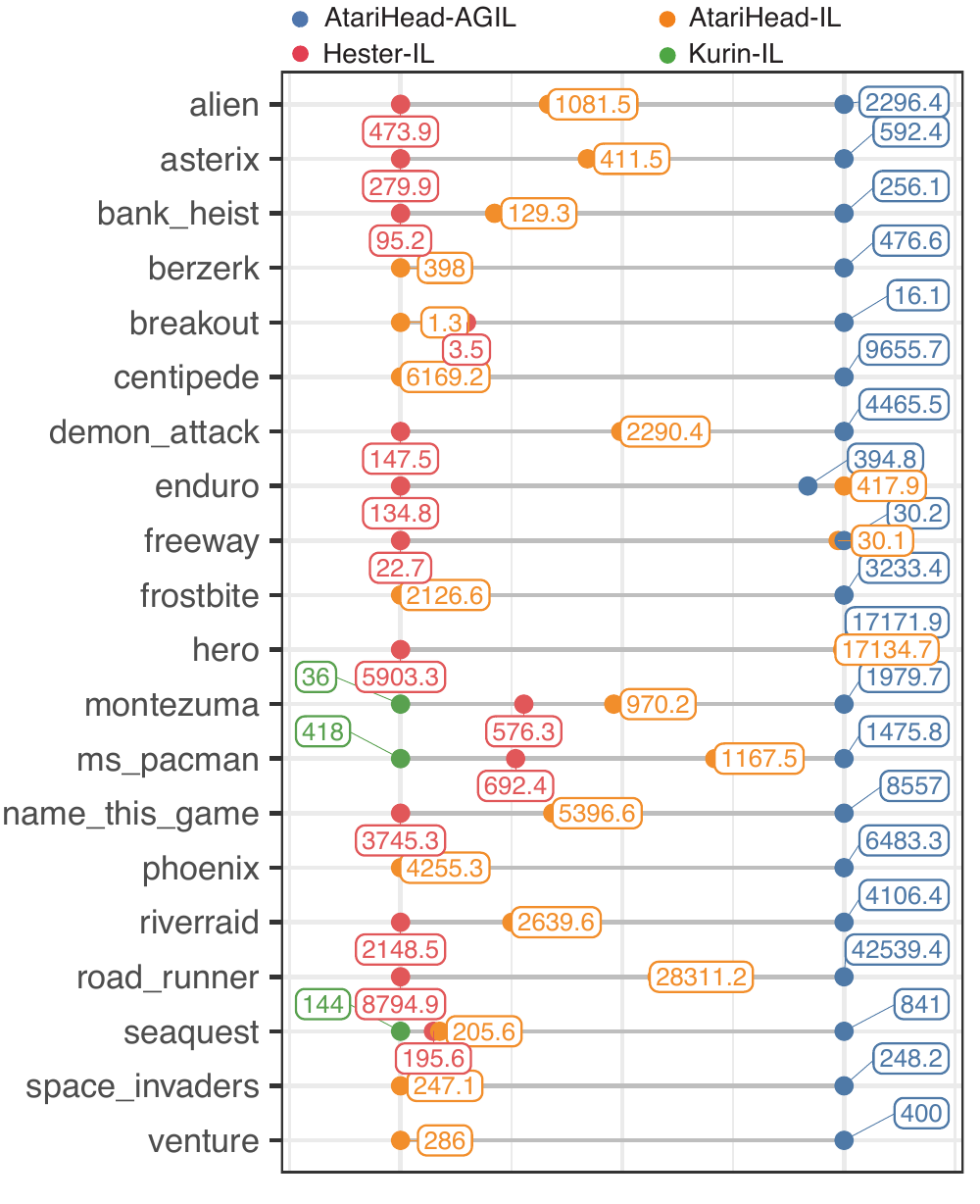}
\caption{Mean game scores of game agents using different data. With the large-scale high-quality dataset we collected, an IL agent is able to perform better than similar agents reported in previous datasets~\cite{hester2018deep,kurin2017atari}. Additionally, incorporating the attention model learned from human gaze improves IL agent's performance with an average improvement of 115.26\%. }
\label{fig:gamescore}
\end{figure}


%% file: 6-discuss.tex
\section{Discussion and Future Work}
We introduce Atari-HEAD, a large-scale dataset of human demonstration playing Atari videos games. The novel features of this dataset include human gaze data, and a semi-frame-by-frame gameplay mode. The latter ensures that states and actions are matched, and allow enough decision time for human players. Software tools to extract image features such as optical flow are published along with the dataset, including a customized video player to visualize data. 

This dataset addresses two major issues in IL and IR research; the first being reproducibility and the second being the need to bridge attention and control. We do this by providing human data that allows researchers to study how humans use visual attention to solve visuomotor tasks. We have shown promising results in saliency prediction and IL using Atari-HEAD. The most exciting result is that the human attention model improves the performance of the IL algorithm. 

There are a number of promising future directions for research that arise from current progress. In this dataset, decision time is also recorded for every action. Such information could help identify difficult states for human players. Presumably these states should be weighted more during learning process, i.e., through state importance measurements~\cite{li2007focus}. Another cue about state importance may come from the immediate reward we recorded for every decision. 

A byproduct of the semi-frame-by-frame gameplay mode is human \emph{option}~\cite{sutton1999between}. We notice that human players often hold a key down until a sub-goal is reached, then release the key and plan for the next sequence of actions. This naturally segments the decision trajectories into temporally extended actions, or options. It has yet to be explored whether a learning agent can learn from this type of human demonstrated options, but results from hierarchical imitation learning~\cite{le2018hierarchical} indicate that this may indeed be possible.

Although deep RL agents have achieved great performance on many tasks, researchers have attempted to understand these agents' behaviors through visualizing feature saliency maps learned by the deep network~\cite{mousavi2016learning,nikulin2019free}. It would be desirable to know whether the agents and humans pay attention to the same visual features. This would be a first step in a general understanding of what information humans use that AIs do not have access to. As progress is made in this direction those principles of attention can form the basis for further improvements in attention guided imitation learning.

%% file: 8-appendix-attached.tex
\section{}
\setcounter{figure}{0}    
\setcounter{table}{0} 

\begin{figure*}
\centering
\subfloat[Alien]{\includegraphics[width=0.2\textwidth]{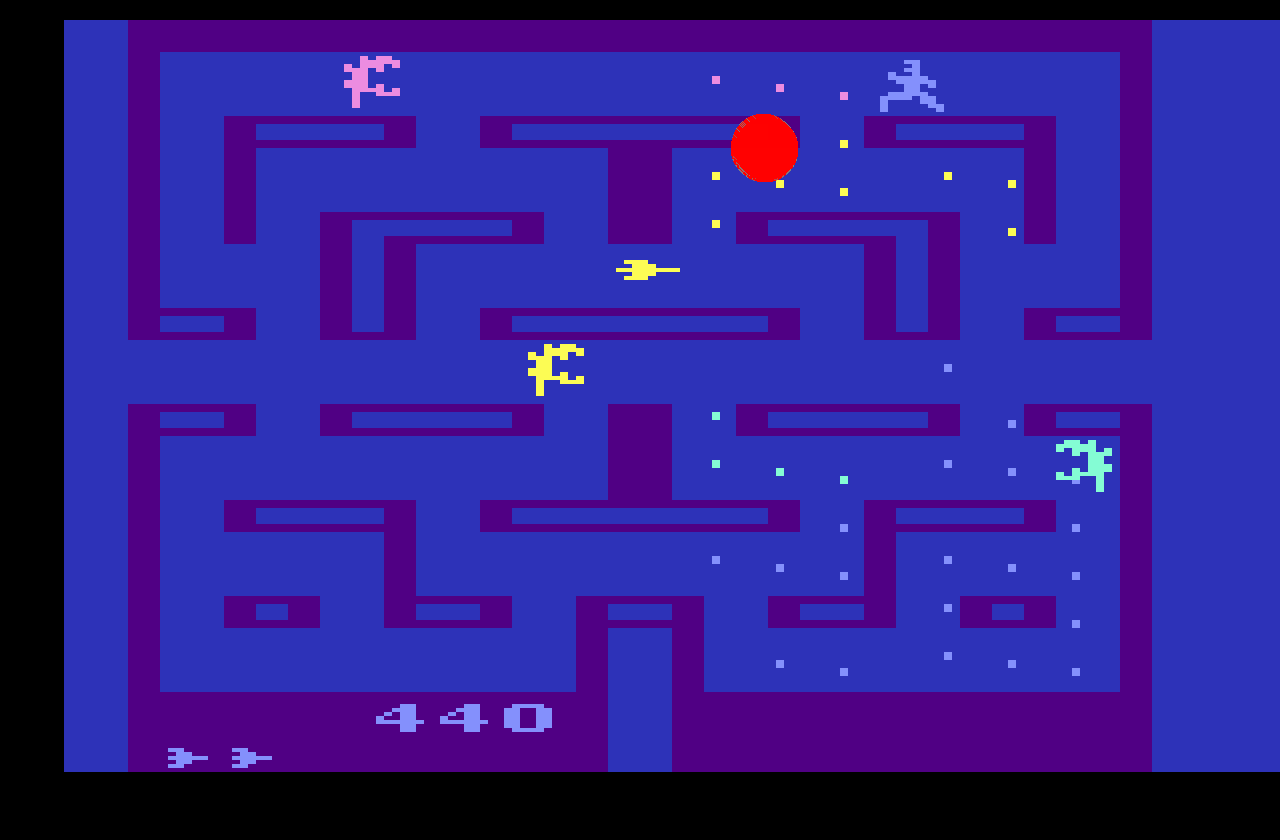}}
\subfloat[Asterix]{\includegraphics[width=0.2\textwidth]{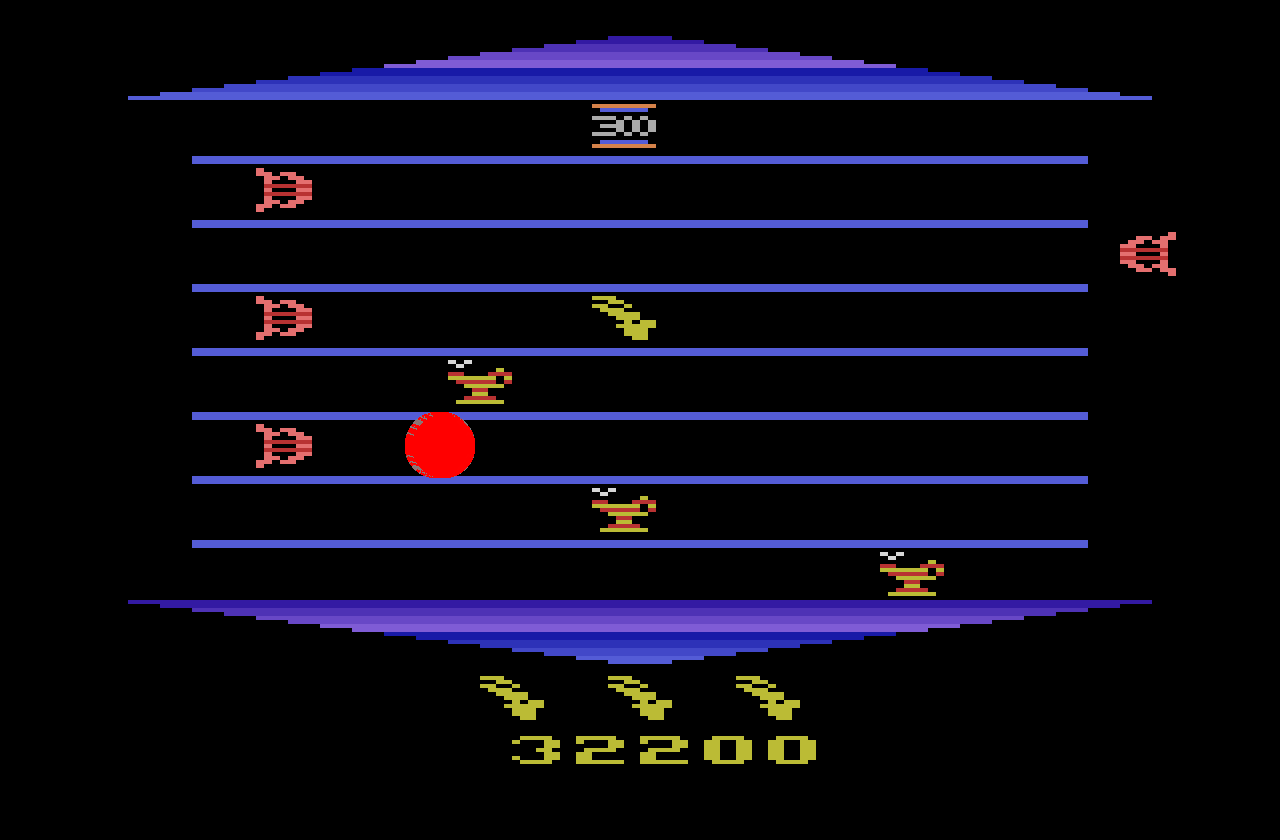}}
\subfloat[Bank Heist]{\includegraphics[width=0.2\textwidth]{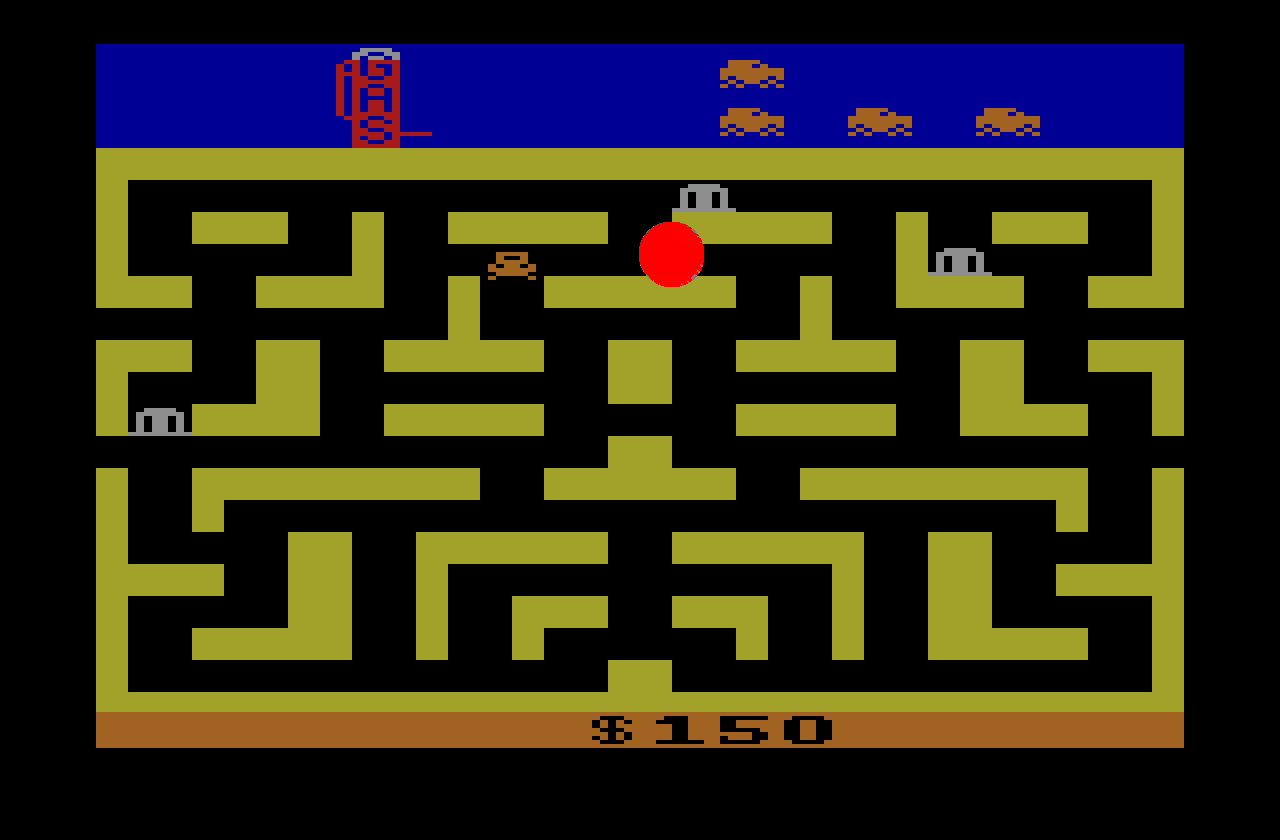}}
\subfloat[Berzerk]{\includegraphics[width=0.2\textwidth]{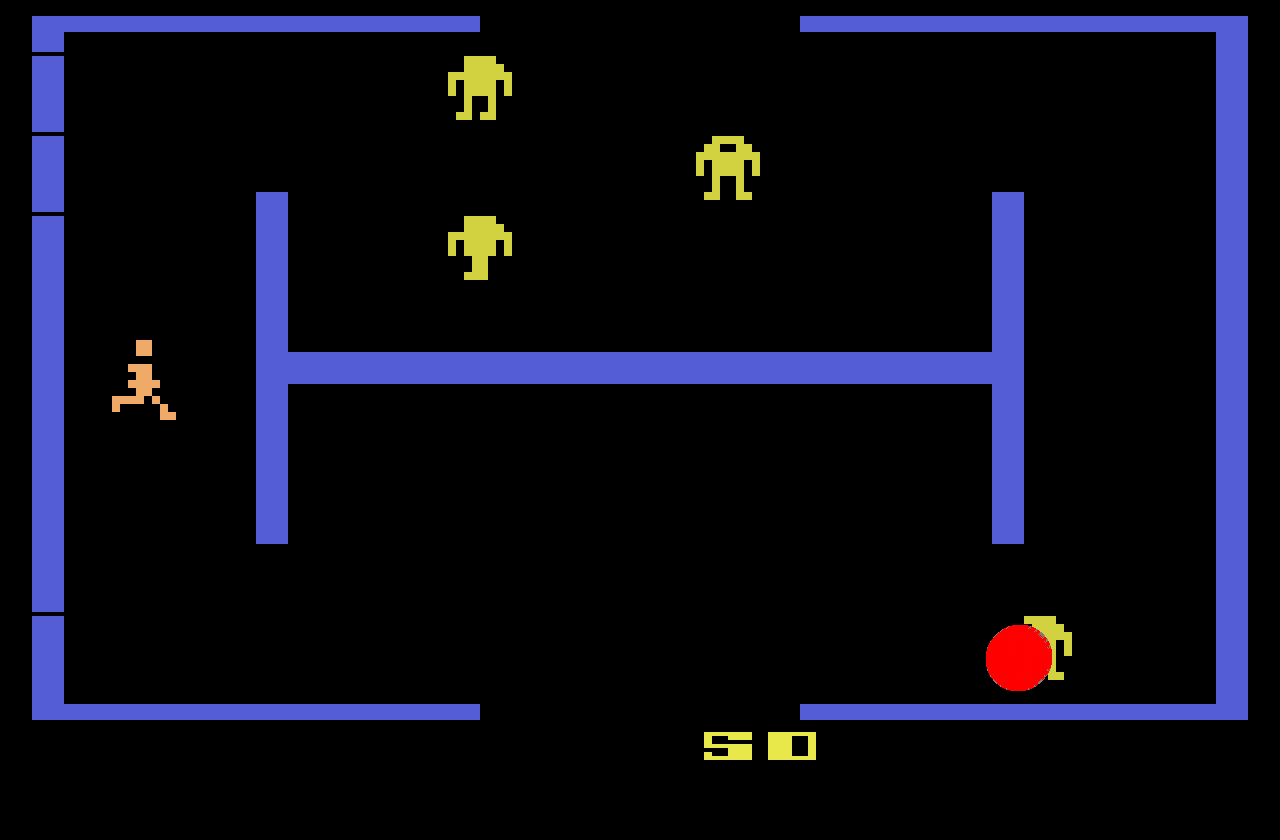}}
\subfloat[Breakout]{\includegraphics[width=0.2\textwidth]{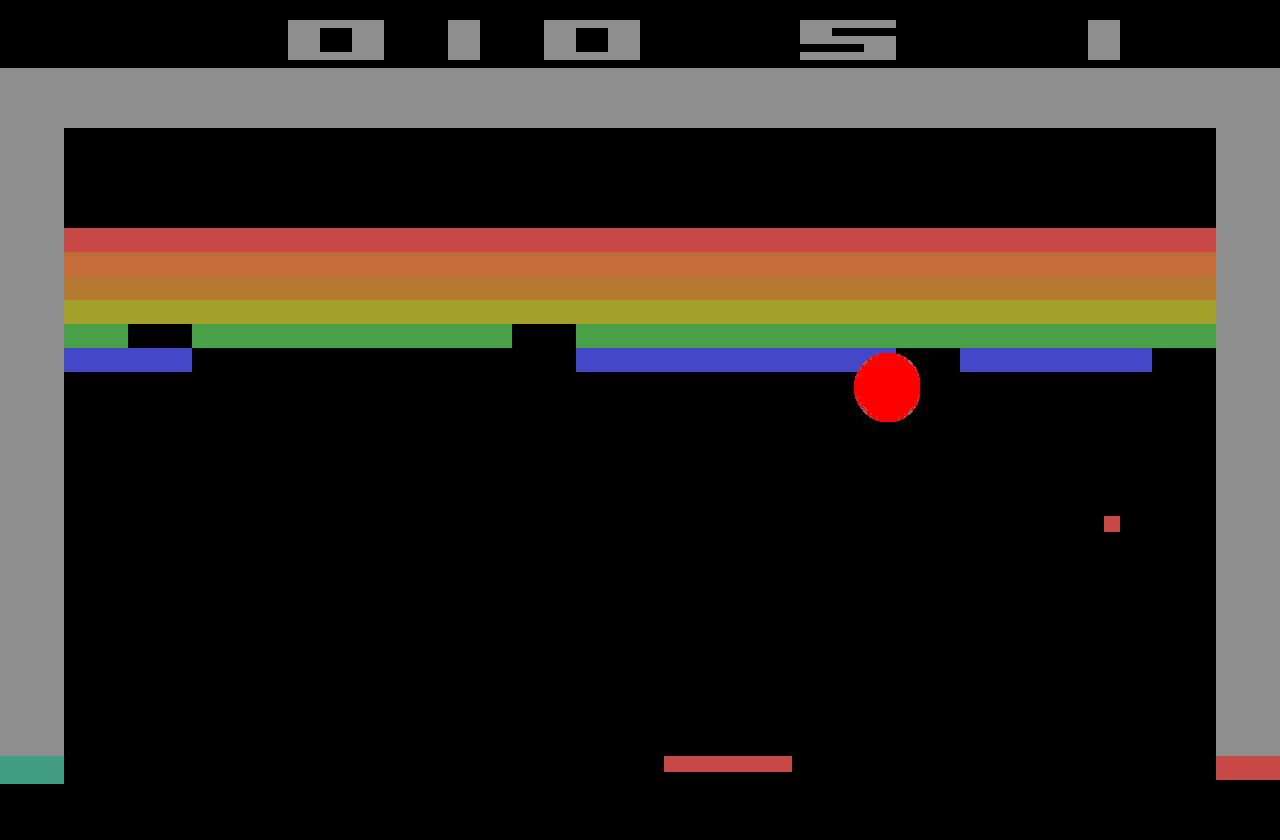}}\\
\subfloat[Centipede]{\includegraphics[width=0.2\textwidth]{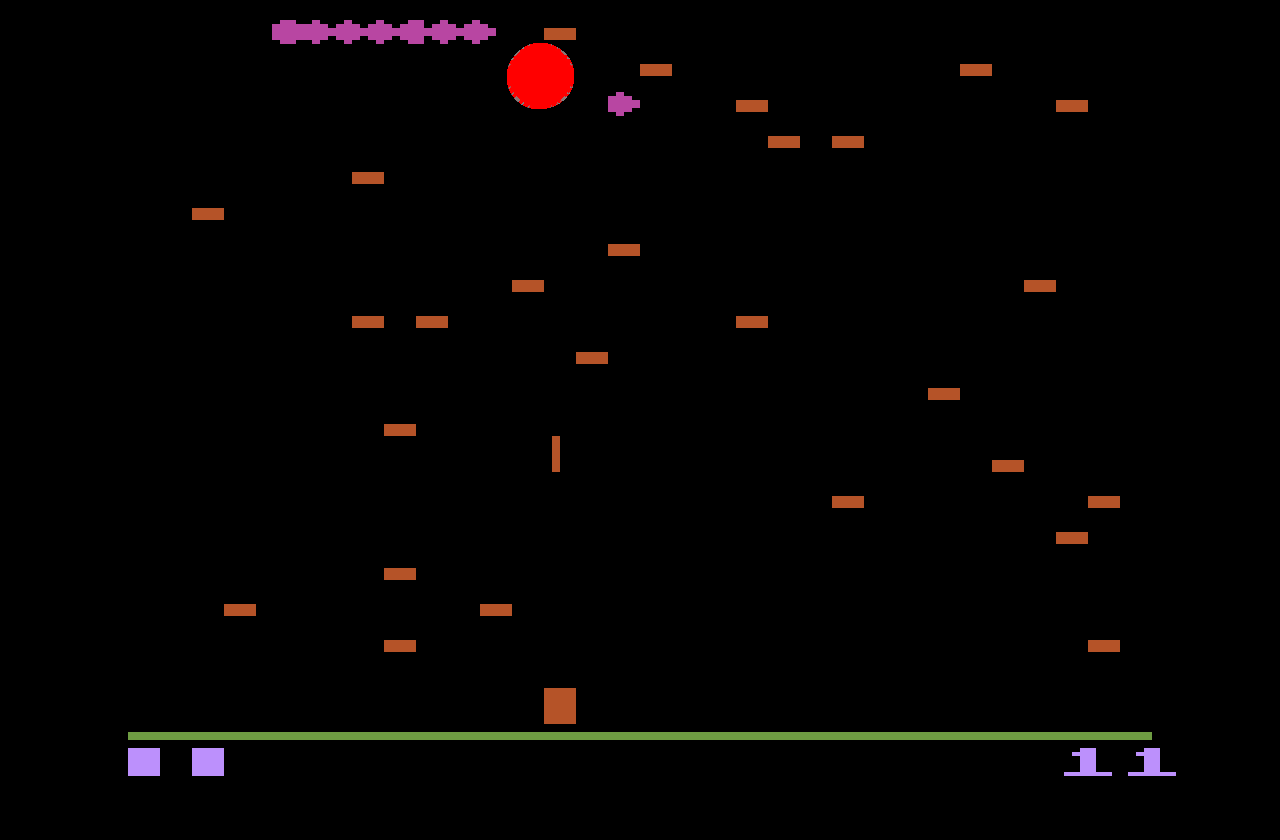}}
\subfloat[Demon Attack]{\includegraphics[width=0.2\textwidth]{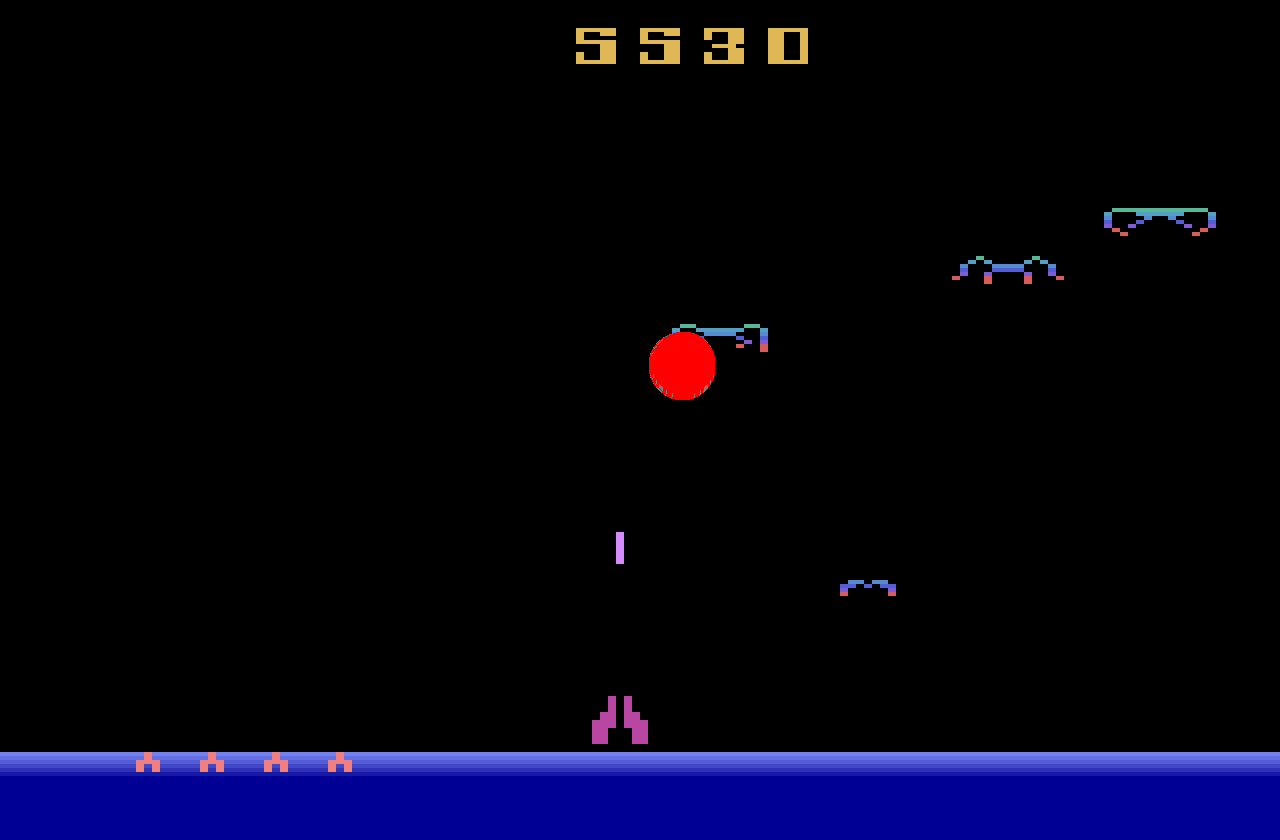}}
\subfloat[Enduro]{\includegraphics[width=0.2\textwidth]{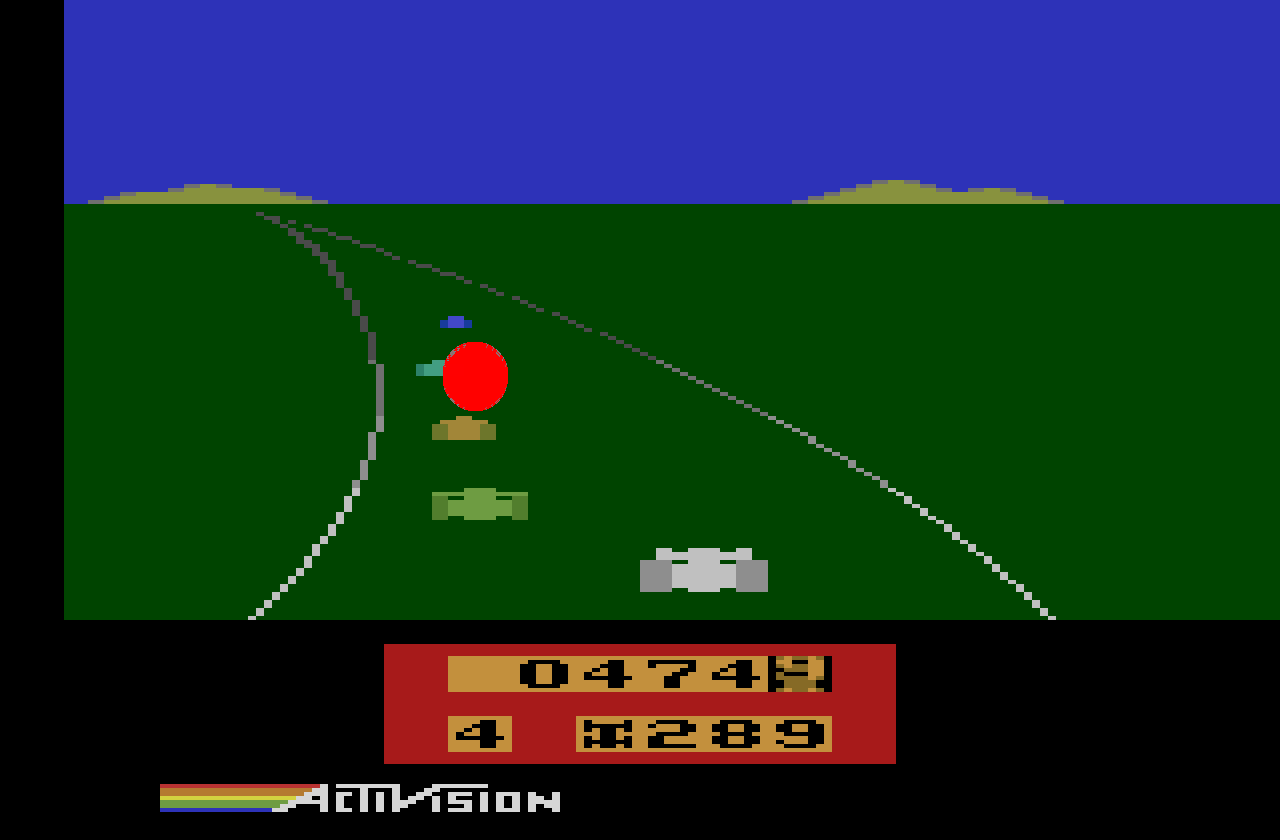}}
\subfloat[Freeway]{\includegraphics[width=0.2\textwidth]{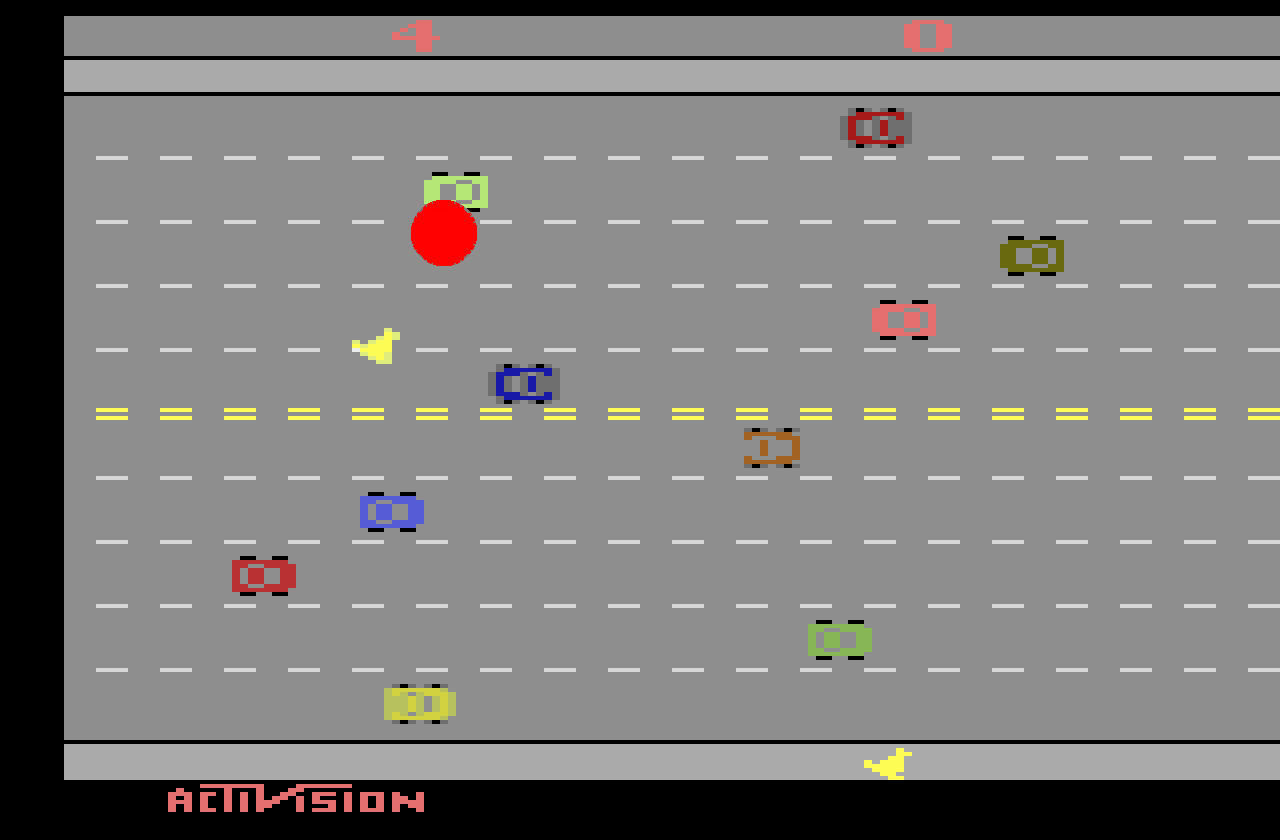}}
\subfloat[Frostbite]{\includegraphics[width=0.2\textwidth]{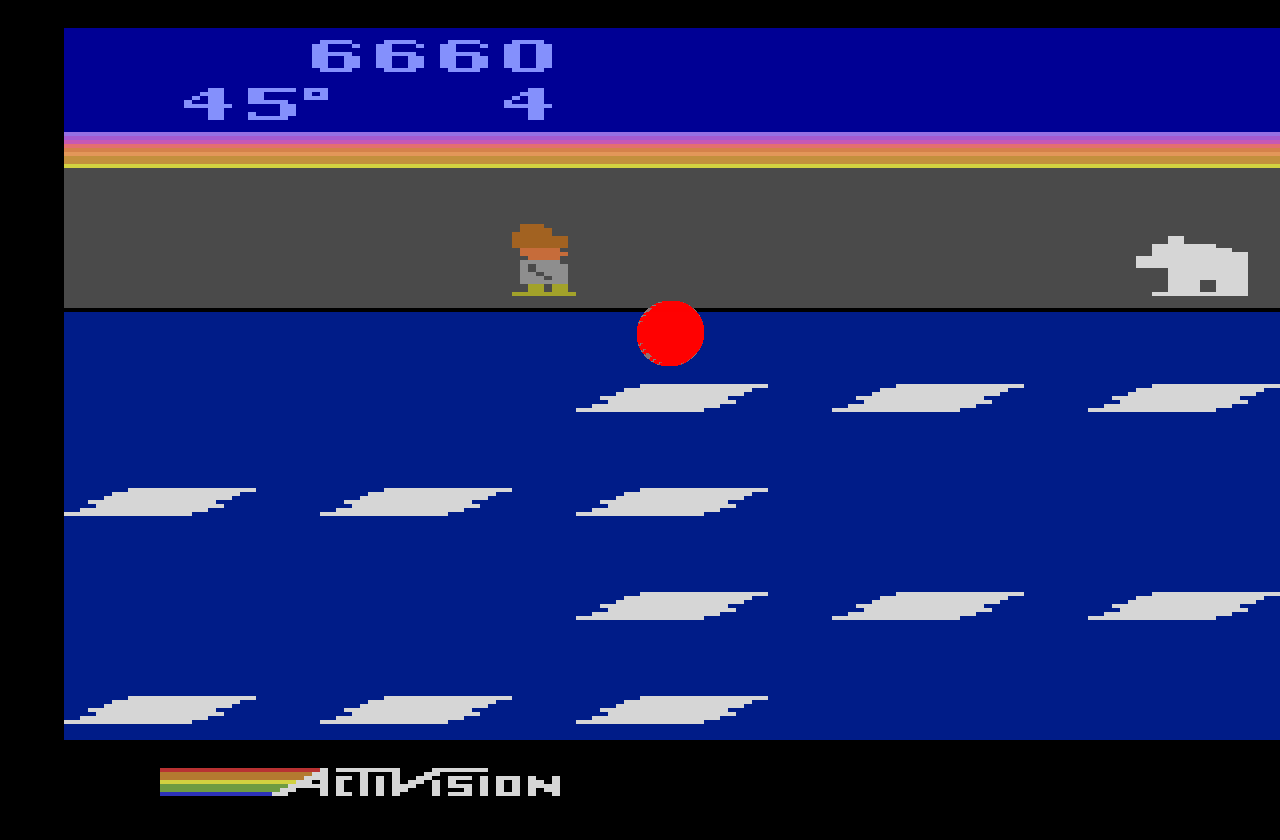}}\\
\subfloat[Hero]{\includegraphics[width=0.2\textwidth]{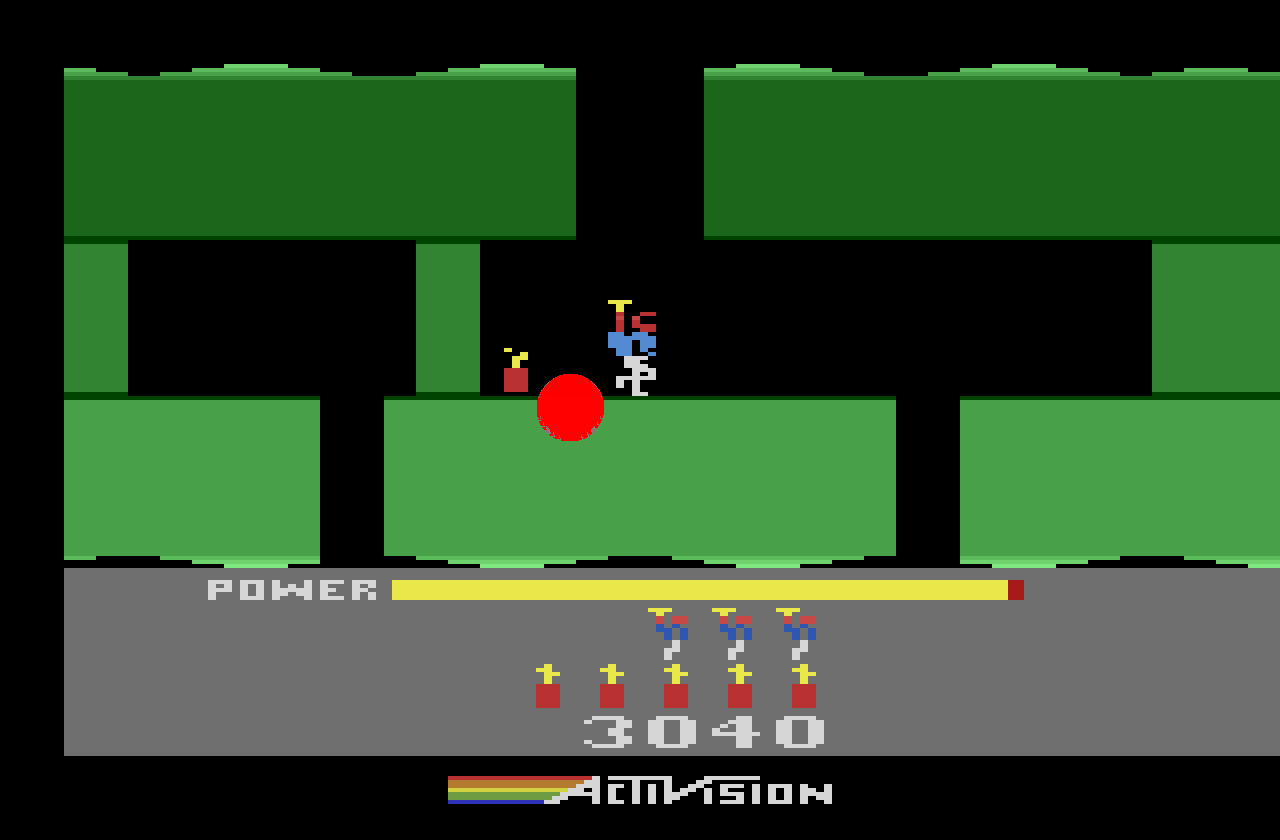}}
\subfloat[Montezuma's Revenge]{\includegraphics[width=0.2\textwidth]{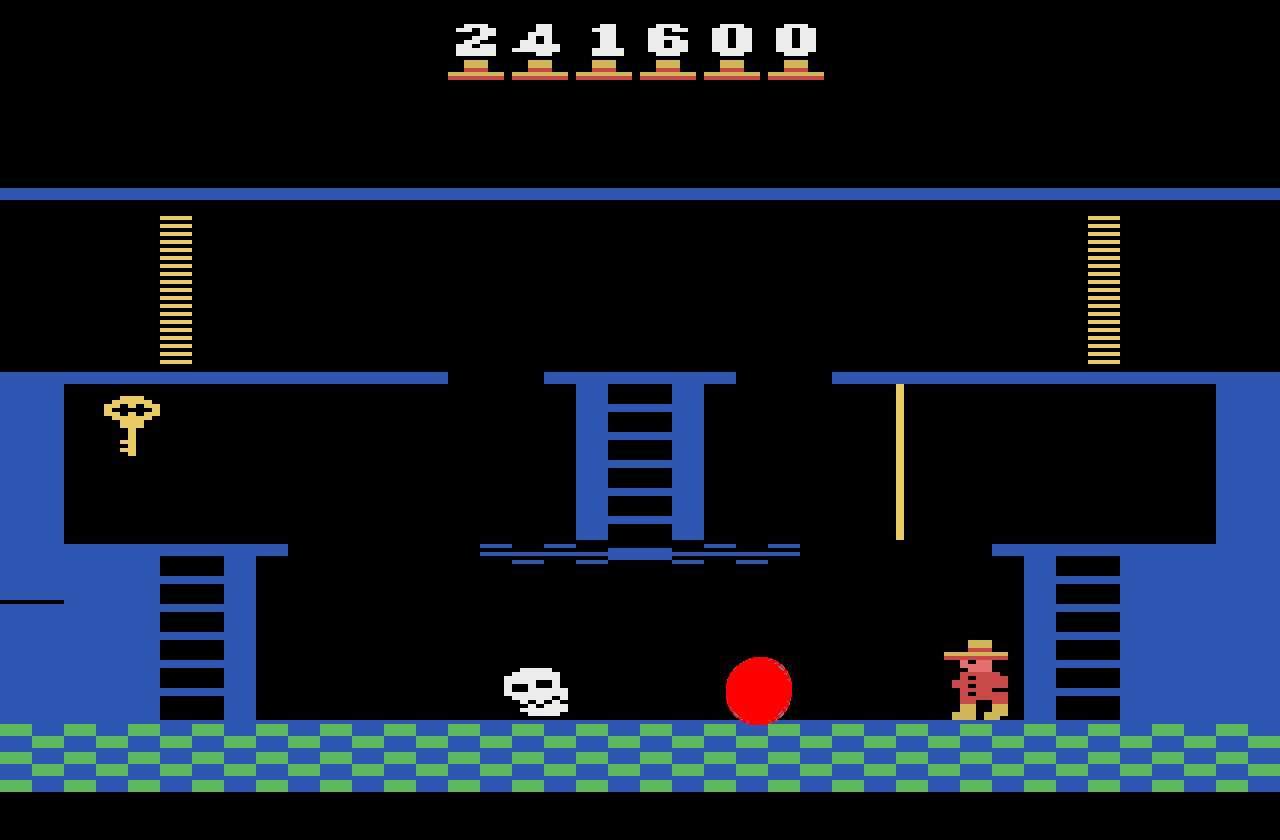}}
\subfloat[Ms.Pacman]{\includegraphics[width=0.2\textwidth]{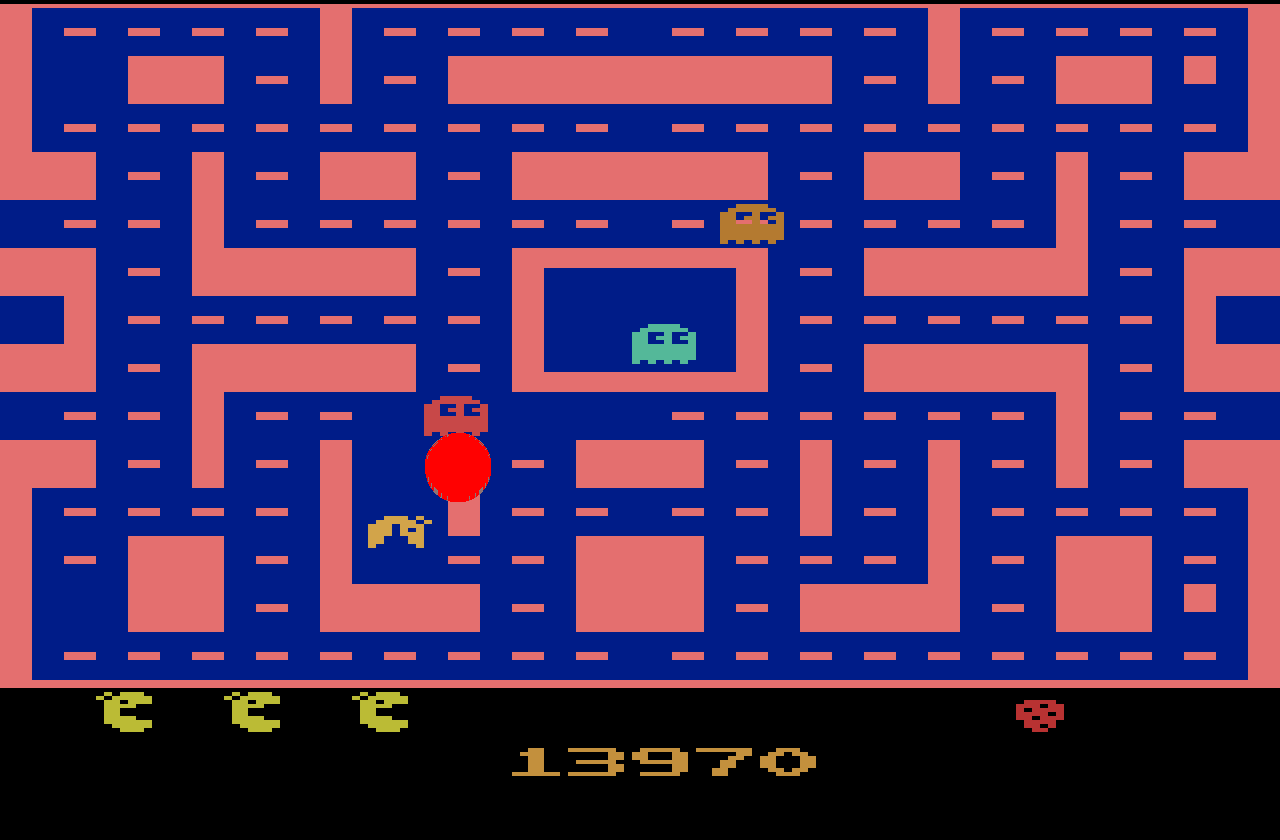}}
\subfloat[Name This Game]{\includegraphics[width=0.2\textwidth]{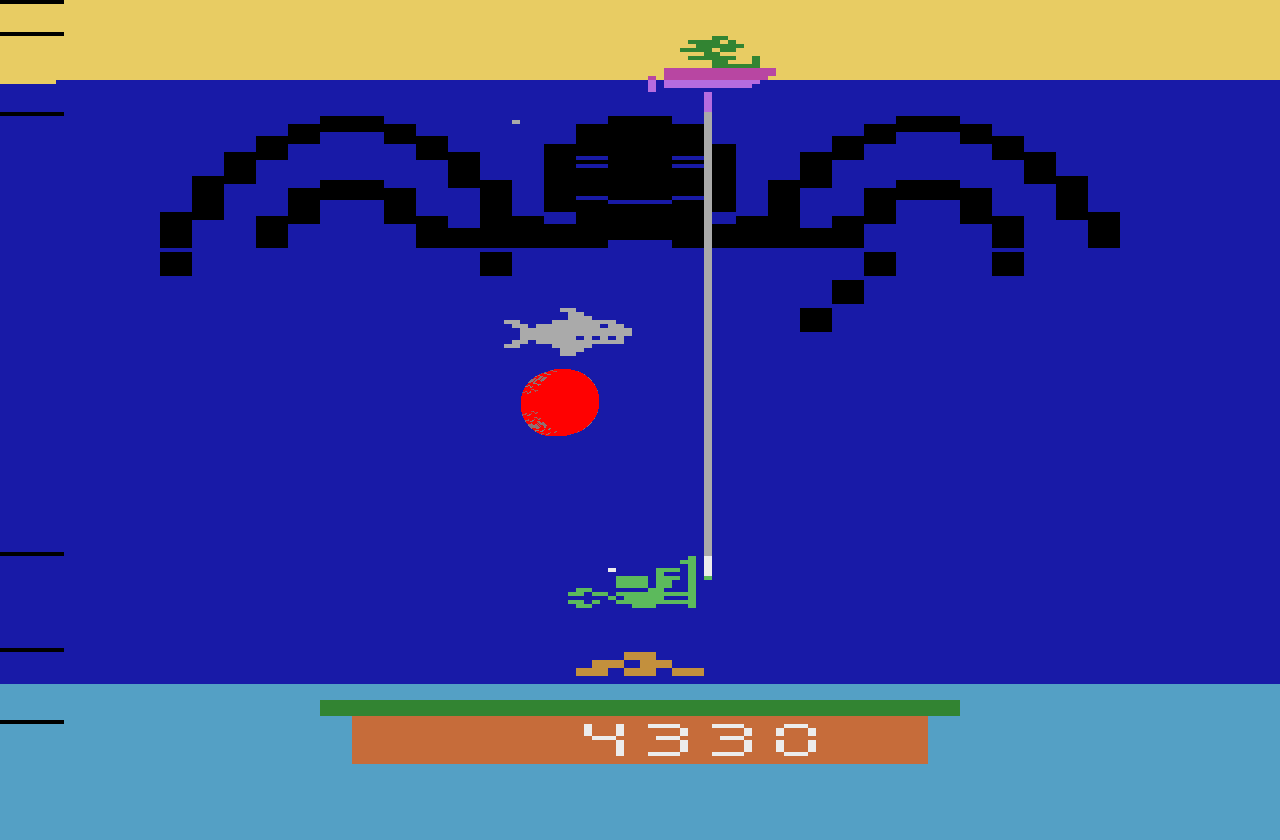}}
\subfloat[Phoenix]{\includegraphics[width=0.2\textwidth]{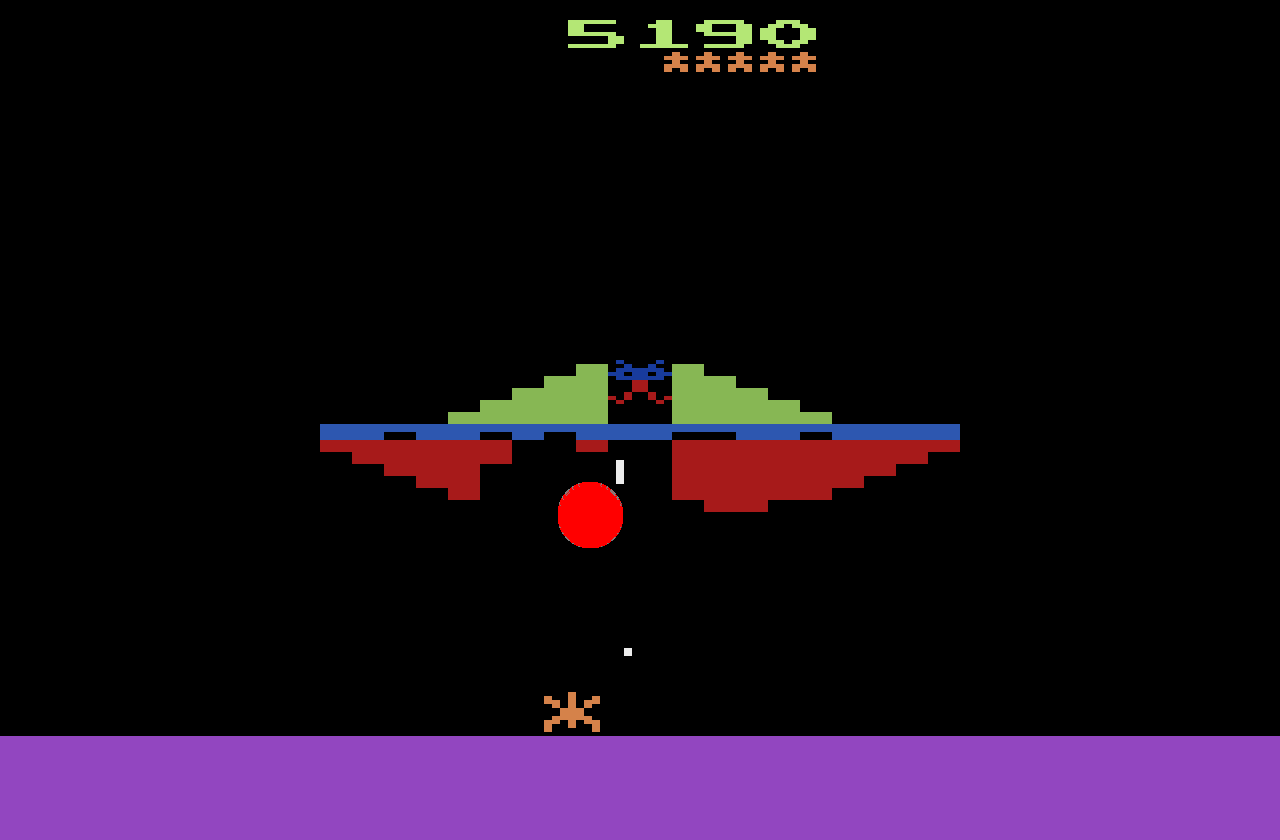}}\\
\subfloat[River Raid]{\includegraphics[width=0.2\textwidth]{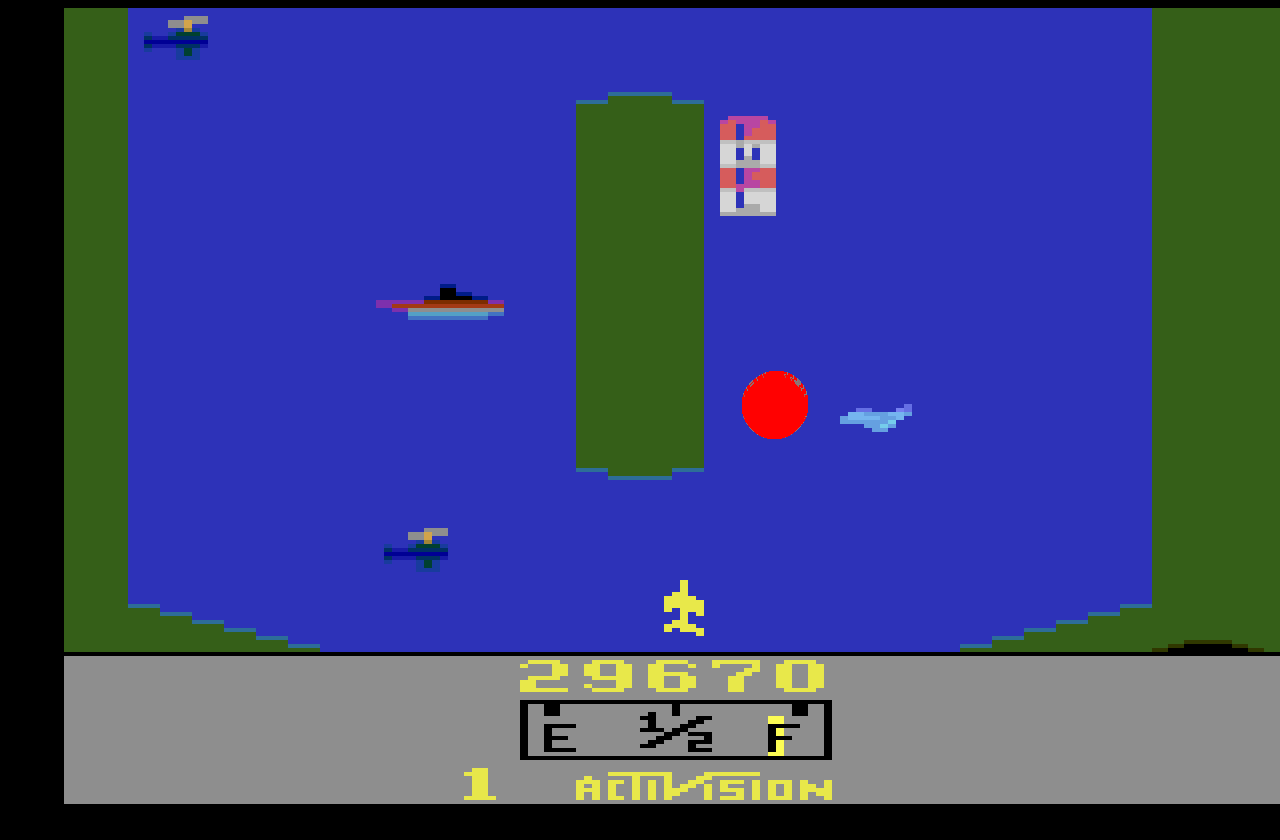}}
\subfloat[Road Runner]{\includegraphics[width=0.2\textwidth]{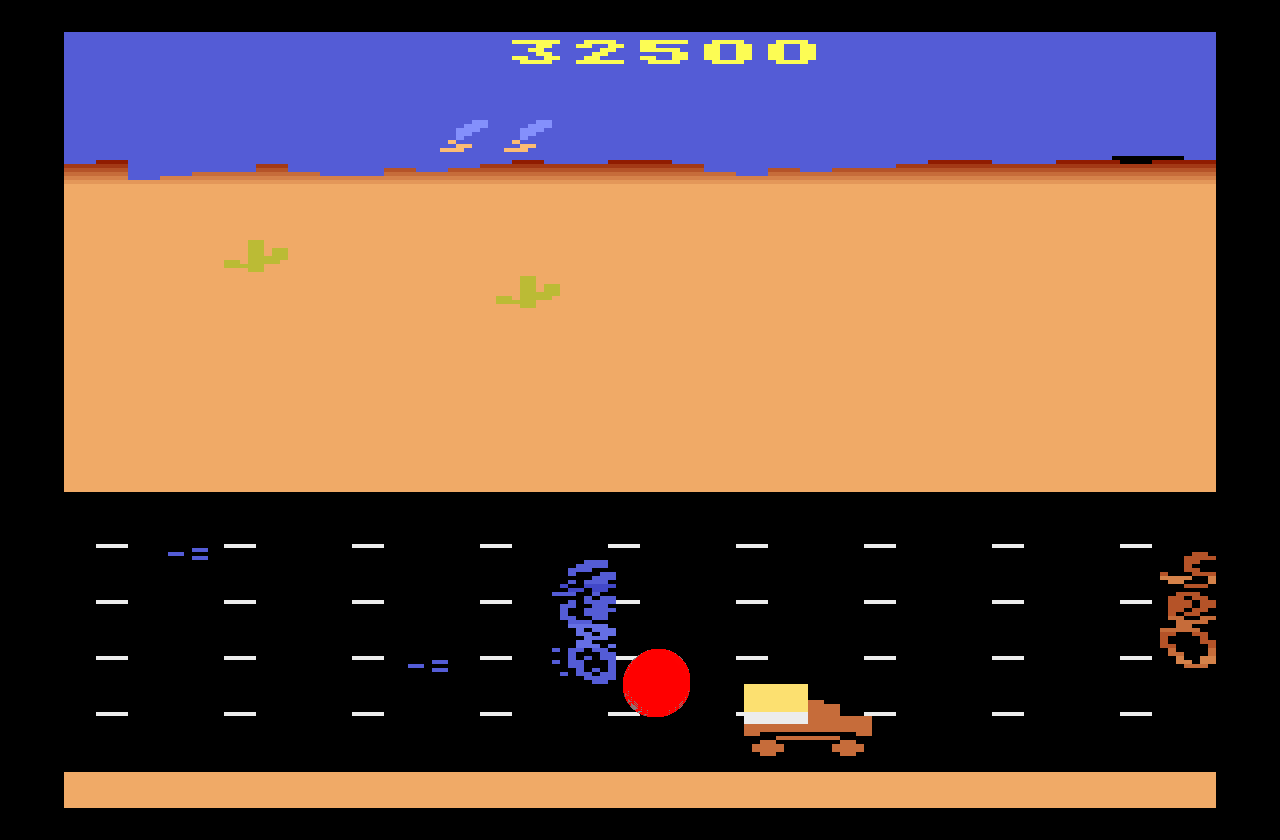}}
\subfloat[Seaquest]{\includegraphics[width=0.2\textwidth]{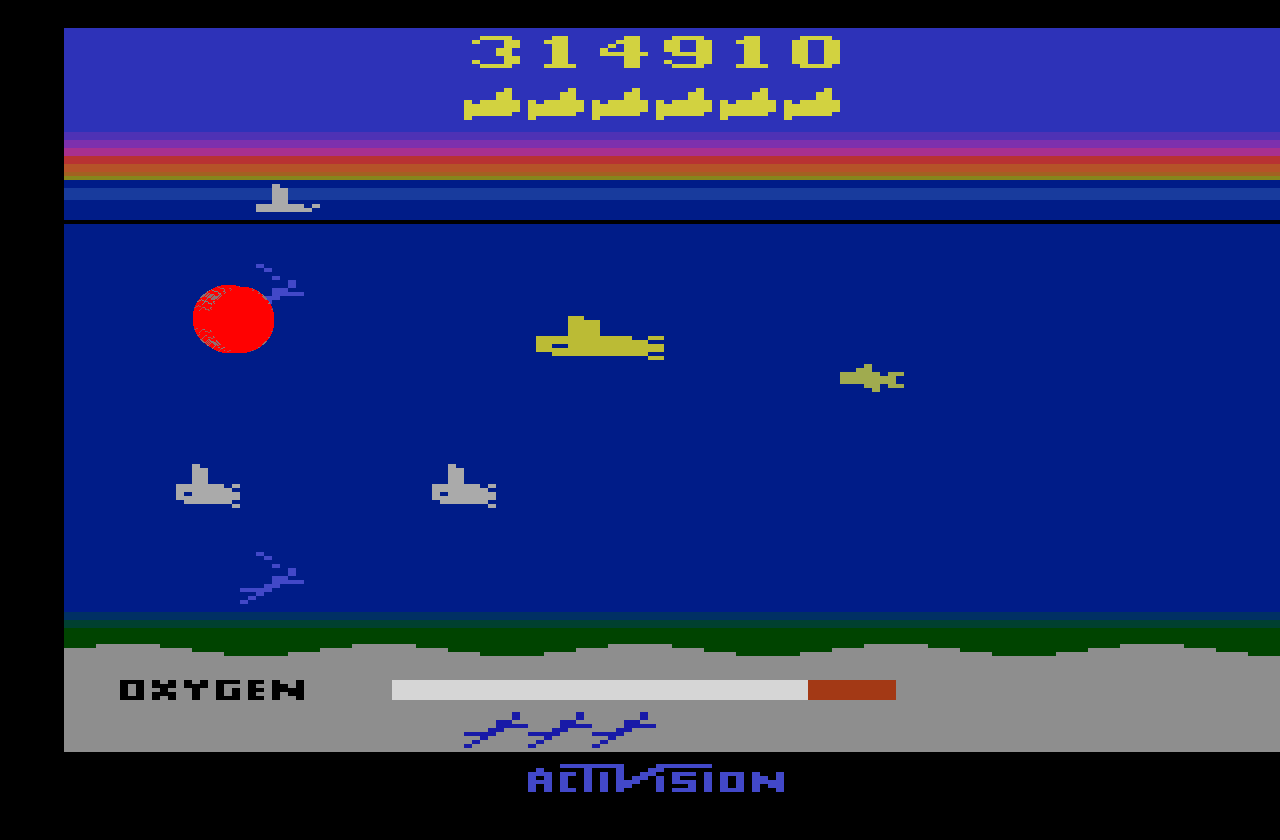}}
\subfloat[Space Invaders]{\includegraphics[width=0.2\textwidth]{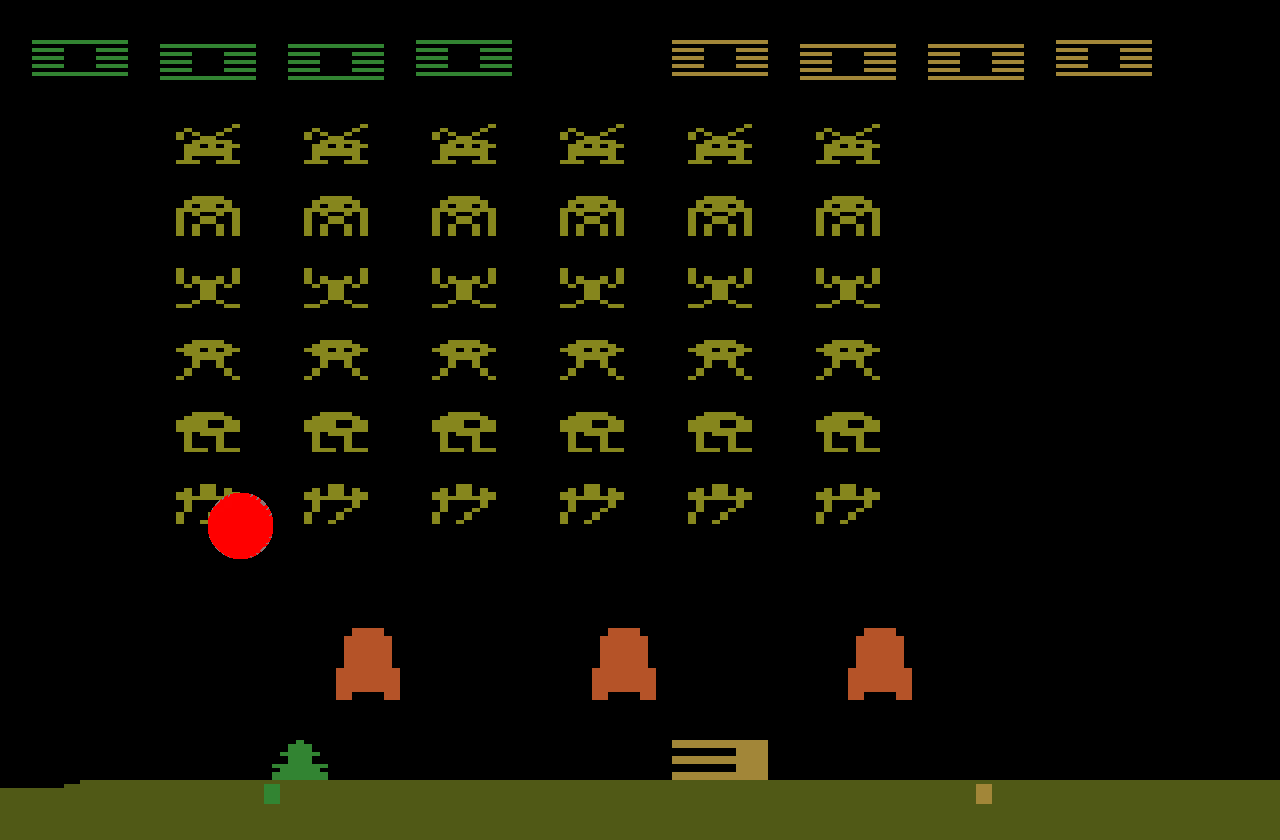}}
\subfloat[Venture]{\includegraphics[width=0.2\textwidth]{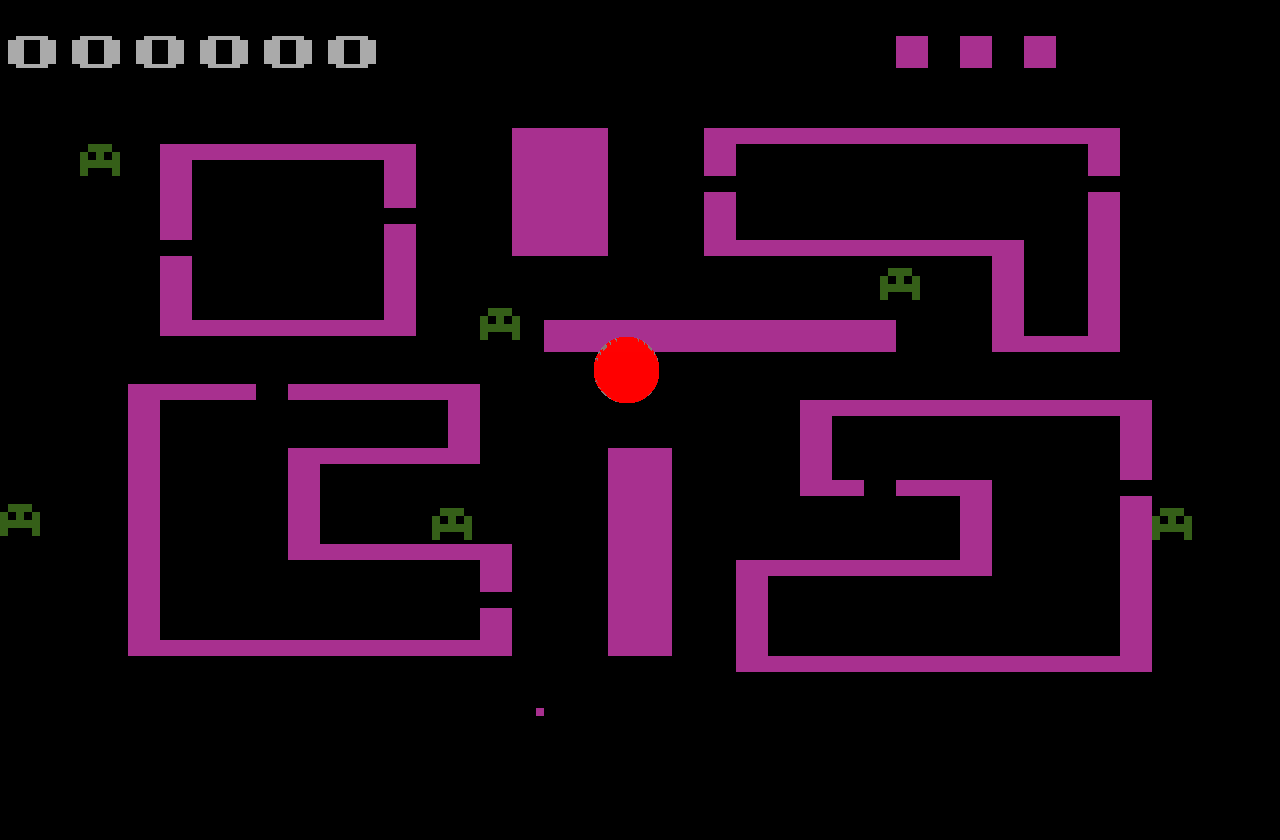}}
\caption{20 Atari 2600 games~\cite{bellemare2012arcade} were used to collect human gaze and action data. Red dot indicates human gaze positions. Atari game platform is a rich environment with games of very different dynamics, visual features, and reward functions. Using these games for
studying visuomotor control is standard in reinforcement and imitation learning. These games capture many interesting aspects of real-world problems, such as the intercepting task in Breakout and Asterix, driving in Enduro, path planning in Alien, Bank Hesit and Ms.Pacman, solving a maze in Hero and Montezuma's Revenge, and a mixture of tasks in Seaquest and Venture.}
\label{fig:gazeresults}
\end{figure*}

\begin{figure*}
\centering
\includegraphics[width=1\textwidth]{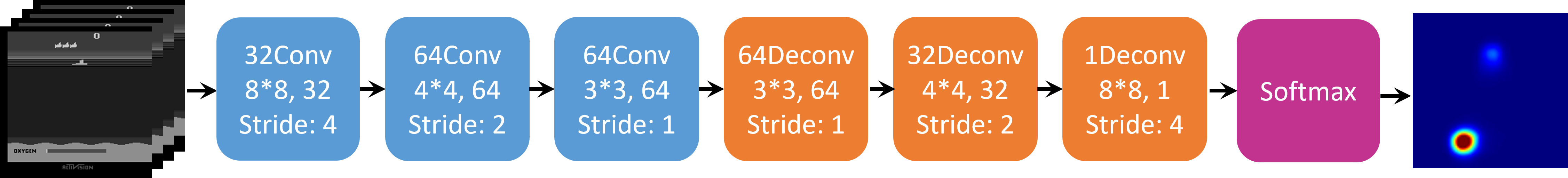}
\caption{The gaze prediction network. The network takes in a stack of 4 consecutive game images in grayscale, passes the inputs to 3 convolutional layers followed by 3 deconvolutional layers. The final output is a gaze saliency map that indicates the predicted probability distribution of the gaze. }
\label{fig:gazenet}
\end{figure*}

\begin{table*}[]
\begin{tabular}{l|llll|llll|llll}
\hline
& \multicolumn{4}{c}{Gaze Network}      & \multicolumn{4}{c}{Bottom-up Saliency}   & \multicolumn{4}{c}{Optical Flow}     \\
& NSS   & AUC   & KL    & CC    & NSS               & AUC   & KL    & CC     & NSS          & AUC   & KL     & CC    \\
\hline
alien              & 6.511 & 0.973 & 1.309 & 0.578 & -0.442            & 0.396 & 4.714 & -0.061 & 1.093        & 0.730 & 8.066  & 0.115 \\
asterix            & 4.846 & 0.966 & 1.556 & 0.485 & 0.104             & 0.526 & 4.166 & 0.001  & 1.330        & 0.711 & 9.959  & 0.151 \\
bank\_heist        & 6.543 & 0.974 & 1.286 & 0.588 & -0.639            & 0.302 & 4.511 & -0.077 & 1.669        & 0.687 & 11.089 & 0.161 \\
berzerk            & 5.280 & 0.966 & 1.530 & 0.503 & 0.834             & 0.630 & 3.903 & 0.077  & 1.523        & 0.646 & 12.955 & 0.163 \\
breakout           & 6.147 & 0.972 & 1.266 & 0.583 & -0.047            & 0.499 & 4.379 & -0.005 & 2.236        & 0.665 & 13.105 & 0.206 \\
centipede          & 5.056 & 0.956 & 1.750 & 0.473 & 0.562             & 0.673 & 3.885 & 0.048  & 1.276        & 0.717 & 11.304 & 0.131 \\
demon\_attack      & 7.662 & 0.980 & 1.084 & 0.645 & -0.247            & 0.576 & 4.835 & -0.034 & 1.752        & 0.764 & 9.672  & 0.178 \\
enduro             & 8.421 & 0.988 & 0.830 & 0.703 & -0.248            & 0.465 & 4.454 & -0.032 & 0.611        & 0.728 & 8.672  & 0.080 \\
freeway            & 7.621 & 0.976 & 1.133 & 0.641 & -0.158            & 0.562 & 4.288 & -0.023 & 1.106        & 0.700 & 10.863 & 0.106 \\
frostbite          & 5.554 & 0.961 & 1.532 & 0.521 & -0.089            & 0.464 & 4.346 & -0.017 & 0.625        & 0.620 & 12.774 & 0.072 \\
hero               & 7.798 & 0.979 & 1.061 & 0.653 & 0.153             & 0.554 & 3.955 & 0.019  & 1.893        & 0.707 & 11.237 & 0.195 \\
montezuma & 8.267 & 0.984 & 0.939 & 0.683 & 0.312             & 0.654 & 3.816 & 0.038  & 1.092        & 0.684 & 12.018 & 0.119 \\
ms\_pacman         & 4.674 & 0.945 & 1.858 & 0.453 & -0.380            & 0.416 & 4.690 & -0.049 & 1.018        & 0.668 & 12.154 & 0.100 \\
name\_this\_game   & 8.164 & 0.977 & 1.111 & 0.653 & -0.559            & 0.367 & 4.855 & -0.069 & 0.831        & 0.609 & 14.039 & 0.086 \\
phoenix            & 7.122 & 0.980 & 1.153 & 0.612 & -0.256            & 0.549 & 4.921 & -0.030 & 1.737        & 0.742 & 10.415 & 0.173 \\
riverraid          & 6.218 & 0.966 & 1.497 & 0.534 & 0.063             & 0.482 & 4.246 & -0.010 & 1.221        & 0.727 & 9.513  & 0.126 \\
road\_runner       & 6.544 & 0.973 & 1.307 & 0.581 & -0.234            & 0.421 & 4.227 & -0.036 & 1.626        & 0.770 & 8.049  & 0.170 \\
seaquest           & 6.350 & 0.964 & 1.469 & 0.552 & -0.258            & 0.345 & 4.799 & -0.042 & 1.725        & 0.742 & 10.147 & 0.171 \\
space\_invaders    & 6.574 & 0.982 & 1.150 & 0.604 & -0.277            & 0.468 & 4.758 & -0.036 & 0.847        & 0.613 & 14.347 & 0.087 \\
venture            & 5.724 & 0.960 & 1.605 & 0.513 & 0.451             & 0.608 & 3.852 & 0.052  & 1.110        & 0.659 & 12.343 & 0.114\\
\hline
\hline
\end{tabular}
\caption{Quantitative results of predicting human gaze across 20 games. Random prediction baseline: NSS = 0.000, AUC = 0.500, KL = 6.100, CC = 0.000. For comparison, the performance of the classic bottom-up saliency~\cite{itti1998model} and optical flow~\cite{farneback2003two} models are also computed. A separate convolution-deconvolution network gaze network (Fig.~\ref{fig:gazenet}) is trained for each individual game. The gaze networks are accurate in predicting human gaze (AUC$>$0.94) for all games. }
\end{table*}

\begin{figure*}
\centering
\subfloat[Ms.Pacman]{\includegraphics[width=1\textwidth]{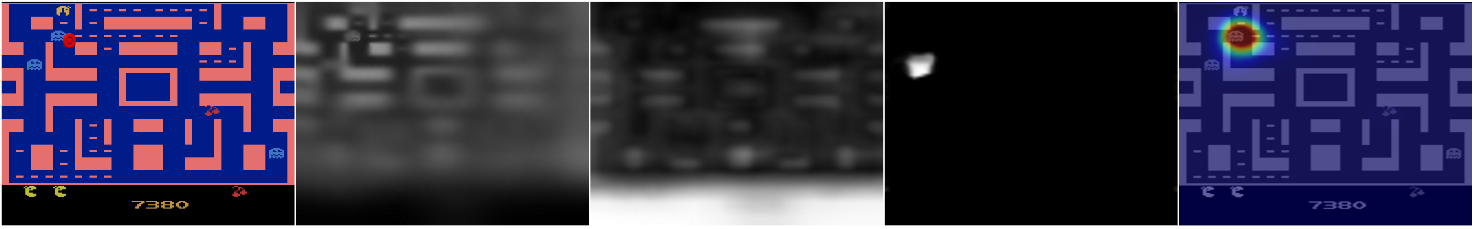}}\\
\subfloat[Seaquest]{\includegraphics[width=1\textwidth]{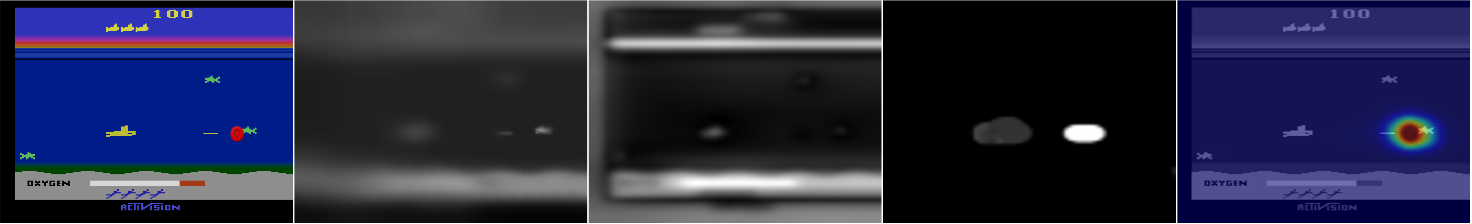}}\\
\subfloat[Space Invaders]{\includegraphics[width=1\textwidth]{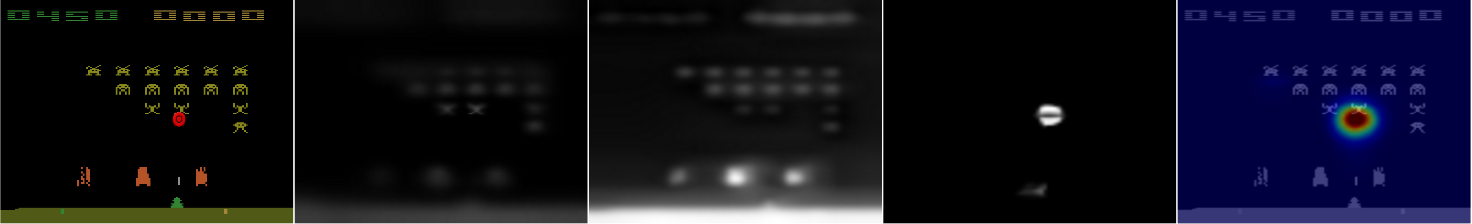}}\\
\caption{Gaze prediction results for 3 games. First column: game screenshots with red dots indicating the human gaze positions. Second column: biologically plausible retinal image, generated by foveated rendering algorithm~\cite{perry2002gaze}. Third column: image saliency calculated by the classic Itti-Koch saliency model~\cite{itti1998model}. Fourth column: Farnebeck optical flow, calculated using the frame in the first column and its previous frame~\cite{farneback2003two}. Fifth column: predicted gaze distribution by convolution-deconvolution network, overlayed on top of the original image. }
\label{fig:gaze2}
\end{figure*}

\begin{figure*}
\centering
\includegraphics[width=0.6\textwidth]{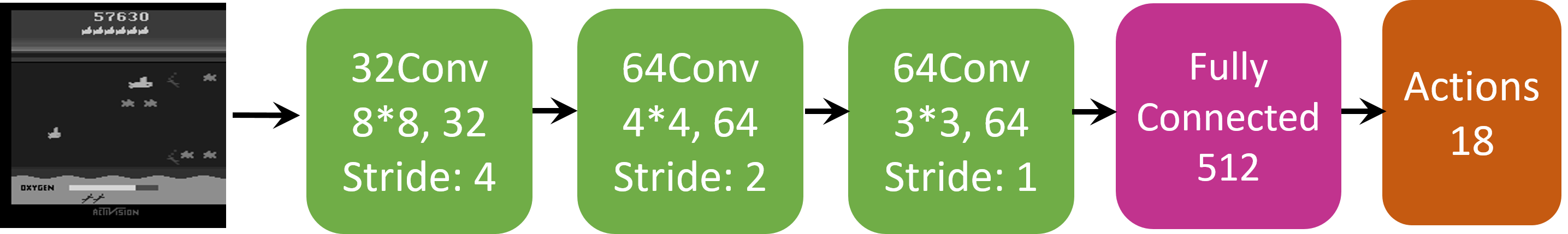}
\caption{ An imitation learning (behavior cloning) network to predict human actions. The network takes in a single grayscale game image as input, and outputs a vector that gives the probability of each action.}
\label{fig:policynet1}
\end{figure*}

\begin{figure*}
\centering
\includegraphics[width=1\textwidth]{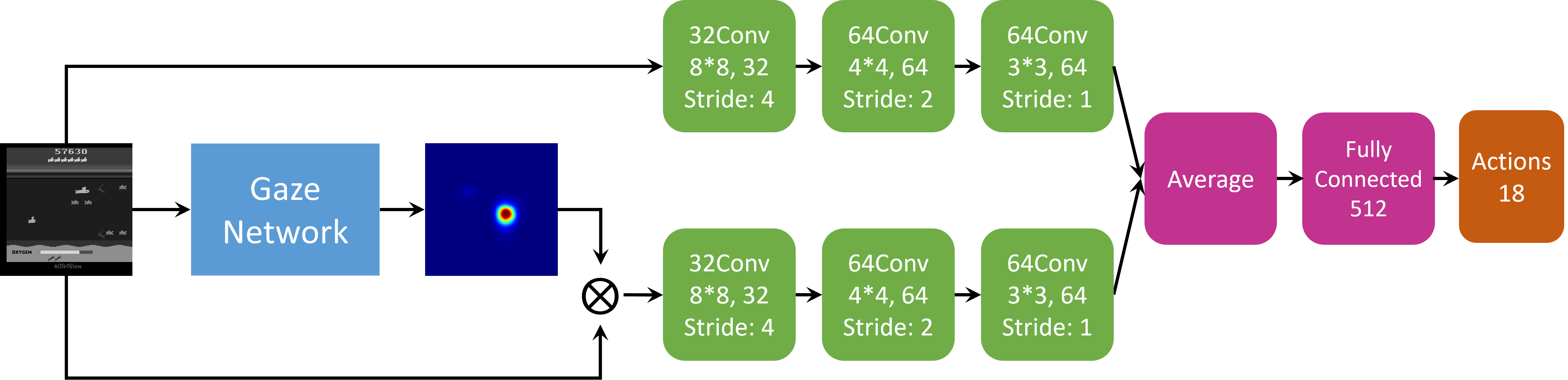}
\caption{The policy network architecture for imitating human actions. The top channel takes in the current image frame and the bottom channel takes in the masked image which is an element-wise product of the original image and predicted gaze saliency map by the gaze network. We then average the output of the two channels.}
\label{fig:policynet2}
\end{figure*}

\begin{table*}
\centering
\scalebox{1}{
\begin{tabular}{c | c c c c }
\hline
Games              & Majority baseline & AtariHead-IL  & AtariHead-AGIL & Improvement            \\
\hline
alien              & 0.293             & 0.504              & 0.690    & +0.185       \\
asterix            & 0.365             & 0.410              & 0.532    & +0.122       \\
bank\_heist        & 0.278             & 0.604              & 0.617    & +0.013       \\
berzerk            & 0.247             & 0.437              & 0.482    & +0.044       \\
breakout           & 0.800             & 0.807              & 0.816    & +0.009       \\
centipede          & 0.581             & 0.587              & 0.628    & +0.042       \\
demon\_attack      & 0.316             & 0.465              & 0.545    & +0.079       \\
enduro             & 0.406             & 0.426              & 0.473    & +0.047       \\
freeway            & 0.781             & 0.959              & 0.963    & +0.003       \\
frostbite          & 0.520             & 0.520              & 0.639    & +0.120       \\
hero               & 0.483             & 0.833              & 0.837    & +0.004       \\
montezuma\_revenge & 0.257             & 0.866              & 0.888    & +0.023       \\
ms\_pacman         & 0.266             & 0.555              & 0.678    & +0.123       \\
name\_this\_game   & 0.361             & 0.551              & 0.746    & +0.195       \\
phoenix            & 0.291             & 0.574              & 0.658    & +0.084       \\
riverraid          & 0.339             & 0.675              & 0.695    & +0.020       \\
road\_runner       & 0.632             & 0.787              & 0.809    & +0.022       \\
seaquest           & 0.208             & 0.414              & 0.574    & +0.160       \\
space\_invaders    & 0.285             & 0.421              & 0.505    & +0.085       \\
venture            & 0.196             & 0.384              & 0.443    & +0.059      \\
\hline
\hline
\end{tabular}}
\caption{Behavior matching accuracy of different models. Majority baseline simply predicts the majority class in that game (the most frequent action). IL: standard imitation learning through behavior cloning. AGIL: policy network that includes saliency map predicted by the gaze network. We also show the improvement of AGIL over standard IL. Random guess prediction accuracy: 0.056. }
\label{tbl:results}
\end{table*}

\begin{table*}
\centering
\scalebox{1}{
\begin{tabular}{c | c c c c c}
\hline
Games & Kurin-IL & Hester-IL & AtariHead-IL              & AtariHead-AGIL               & Improvement            \\
\hline
alien              & -                 & 473.9   & 1081.5 $\pm$ 741.8   & 2296.4 $\pm$ 1105.7   & +112.33\%    \\
asterix            & -                 & 279.9   & 411.5 $\pm$ 192.6    & 592.4 $\pm$ 290.5     & +43.96\%     \\
bank\_heist        & -                 & 95.2    & 129.3 $\pm$ 75.8     & 256.1 $\pm$ 116.8     & +98.07\%     \\
berzerk            & -                 & -       & 398.0 $\pm$ 189.4    & 476.6 $\pm$ 197.4     & +19.75\%     \\
breakout           & -                 & 3.5     & 1.3 $\pm$ 1.4        & 16.1 $\pm$ 22.5       & +1138.46\%   \\
centipede          & -                 & -       & 6169.2 $\pm$ 3856.1  & 9655.7 $\pm$ 5782.8   & +56.51\%     \\
demon\_attack      & -                 & 147.5   & 2290.4 $\pm$ 1806.7  & 4465.5 $\pm$ 2603.6   & +94.97\%     \\
enduro             & -                 & 134.8   & 417.9 $\pm$ 91.4     & 394.8 $\pm$ 71.2      & -5.53\%     \\
freeway            & -                 & 22.7    & 30.1 $\pm$ 1.2       & 30.2 $\pm$ 1.0        & +0.33\%      \\
frostbite          & -                 & -       & 2126.6 $\pm$ 1444.3  & 3233.4 $\pm$ 1857.5   & +52.05\%     \\
hero               & -                 & 5903.3  & 17134.7 $\pm$ 6454.5 & 17171.9 $\pm$ 8939.8    & +0.22\%     \\
montezuma          & 36 $\pm$ 8.0        & 576.3   & 970.2 $\pm$ 896.2    & 1979.7 $\pm$ 1291.7   & +104.05\%    \\
ms\_pacman         & 418 $\pm$ 20.0      & 692.4   & 1167.5 $\pm$ 686.9   & 1475.8 $\pm$ 858.5    & +26.41\%     \\
name\_this\_game   & -                 & -       & 5396.6 $\pm$ 1757.0  & 8557.0 $\pm$ 2015.6   & +58.56\%     \\
phoenix            & -                 & 3745.3  & 4255.3 $\pm$ 1967.8  & 6483.3 $\pm$ 3051.5   & +52.36\%     \\
riverraid          & -                 & 2148.5  & 2639.6 $\pm$ 669.3   & 4106.4 $\pm$ 1457.1   & +55.57\%     \\
road\_runner       & -                 & 8794.9  & 28311.2 $\pm$ 7261.8 & 42539.4 $\pm$ 11177.2 & +50.26\%     \\
seaquest           & 144 $\pm$ 12.4       & 195.6   & 205.6 $\pm$ 103.7    & 841.0 $\pm$ 842.1     & +309.05\%    \\
space\_invaders    & -                 & -       & 247.1 $\pm$ 149.2    & 248.2 $\pm$ 147.1     & +0.45\%      \\
venture            & -                 & -       & 286.0 $\pm$ 146.8    & 400.0 $\pm$ 175.4     & +39.86\%     \\
\hline
\hline
\end{tabular}}
\caption{Game scores (mean $\pm$ standard deviation) of game agents using different data. Kurin-IL and Hester-IL are imitation learning results reported in~\cite{kurin2017atari} and~\cite{hester2018deep}.  Applying IL and AGIL~\cite{zhang2018agil} to our dataset, the mean scores are averaged over 500 episodes per game, with each episode initialized with a randomly generated seed. The game is cutoff after 108K frames~\cite{hessel2018rainbow}. The agent chooses an action $a$ probabilistically using a softmax function with Gibbs (Boltzmann) distribution according to policy network's prediction $P(a)$:
$\pi(a) = \frac{\exp(\eta P(a))}{\sum_{a' \in \mathcal{A}} \exp(\eta P(a'))}$
where $\mathcal{A}$ denotes the set of all possible actions, $\exp(.)$ denotes the exponential function, and the temperature parameter $\eta$ is set to 1. The scale and quality of our data leads to better performance, when comparing to AtariHEAD-IL to Kurin-IL and Hester-IL. The AtariHead-AGIL agent first learns to predict human gaze and uses the learned gaze model to guide the process of learning human decisions. Incorporating attention leads to an average improvement of 115.26\% over a standard IL algorithm using our dataset.}
\end{table*}